\documentclass[10pt,twocolumn,letterpaper]{article}
\ifdefined\pdfminorversion\pdfminorversion=7\fi
\usepackage[pagenumbers]{cvpr}
\usepackage{multirow}
\usepackage{algorithm}
\usepackage{algorithmic}
\usepackage{placeins}
\definecolor{cvprblue}{rgb}{0.21,0.49,0.74}
\usepackage[breaklinks,colorlinks,hyperfootnotes=false,allcolors=cvprblue]{hyperref}
\def\confName{CVPR}
\def\confYear{2027}
\title{PhysAgent: A Multi-Agent Framework for Reliable Remote Heart Rate Estimation}
\author{%
Yehui Yang\textsuperscript{*1,2}\quad Bo Zhao\textsuperscript{*1}\quad Junzhe Cao\textsuperscript{1}\quad Hui Ma\textsuperscript{1}\\[3pt]
Yue Sun\textsuperscript{3}\quad Wenjin Wang\textsuperscript{4}\quad Zitong Yu\textsuperscript{\textdagger1}\\[7pt]
{\normalsize $^{1}$\,Great Bay University}\\[2pt]
{\normalsize $^{2}$\,Shenzhen University of Advanced Technology}\\[2pt]
{\normalsize $^{3}$\,Macao Polytechnic University}\\[2pt]
{\normalsize $^{4}$\,Southern University of Science and Technology}%
}

\hypersetup{
  pdftitle={PhysAgent: A Multi-Agent Framework for Reliable Remote Heart Rate Estimation},
  pdfauthor={Yehui Yang, Bo Zhao, Junzhe Cao, Hui Ma, Yue Sun, Wenjin Wang, Zitong Yu}
}
\DeclareRobustCommand{\reviewchange}[1]{#1}
\begin{document}
\raggedbottom
\maketitle
\begingroup
\renewcommand{\thefootnote}{\fnsymbol{footnote}}
\footnotetext[1]{Equal contribution.}
\footnotetext[2]{Corresponding author.}
\endgroup

\begin{abstract}
Remote photoplethysmography (rPPG) estimates heart rate from facial videos, but its weak physiological signal is susceptible to motion, illumination changes, and other disturbances. Different models can produce conflicting yet individually plausible heart-rate candidates, making direct fusion prone to mixing incompatible physiological hypotheses. We propose PhysAgent, an inference-time multi-agent candidate-verification framework that requires no retraining of the base models. PhysAgent treats candidate heart rates and their associated waveforms as physiological hypotheses. A multimodal large language model (MLLM)-driven reasoning module integrates visual context, signal quality, and candidate disagreement to generate structured fusion decisions. We further introduce Proposal-Conditioned Hypothesis Verification (PCHV), a constrained fusion method that checks execution validity, candidate-cluster consistency, and detected harmonic relations before numerical execution. Through acceptance, repair, or fallback, PCHV constrains deterministic fusion to a single candidate-hypothesis cluster. Experiments on four public rPPG datasets show that PhysAgent achieves the lowest average mean absolute error (MAE) among the compared methods, reaching 3.60 bpm and 4.83 bpm under intra-dataset and cross-dataset evaluation, respectively.
\end{abstract}

\vspace{-12pt}
\section{Introduction}
\begin{figure*}[t]
\centering
\includegraphics[width=0.98\textwidth]{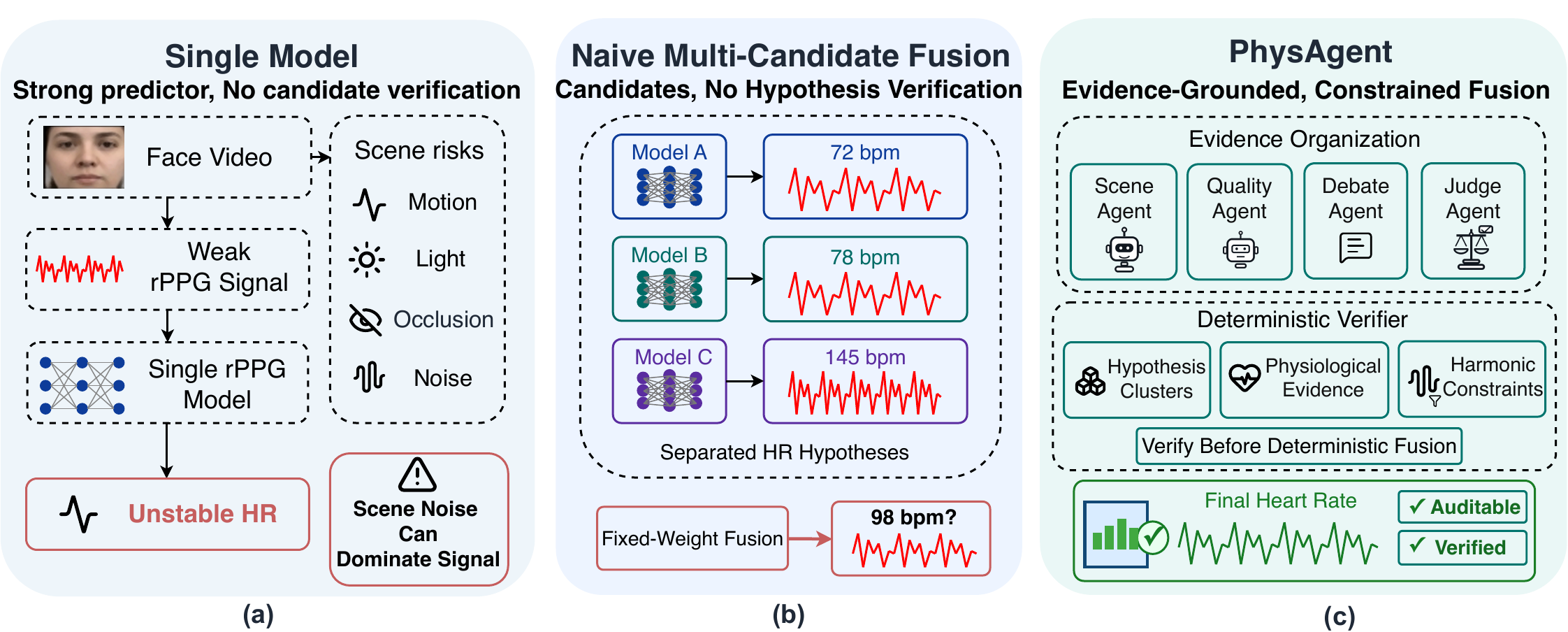}
\caption{(a) A single model can fail when scene-specific nuisance factors overwhelm the weak physiological signal. (b) Naive multi-candidate fusion may combine incompatible heart-rate hypotheses and produce an unsupported estimate. (c) PhysAgent treats base-model outputs as physiological hypotheses, uses MLLM-driven agents to assess candidate reliability and guide selection, and constrains numerical fusion using candidate-cluster and harmonic evidence.}
\label{fig:motivation}
\end{figure*}

Remote photoplethysmography (rPPG) estimates heart rate by recovering subtle skin-color variations induced by cardiac pulsation from ordinary facial videos. Despite substantial advances in deep rPPG estimation \citep{yu2022physformer,li2023motionrobust,zou2025rhythmmamba}, most methods still rely on a single model to predict heart rate or recover pulse waveforms. However, the weak pulse signal can be overwhelmed by motion, illumination changes, and other disturbances, causing a single model to produce unstable heart-rate estimates under challenging video conditions, as illustrated in Figure~\ref{fig:motivation}(a) \citep{shao2025realworld}.

Different rPPG models provide multiple candidate physiological hypotheses for the same video, but these candidates can conflict, and their reliability varies with video conditions, training domain, and signal quality \citep{liu2023efficientphys,li2023motionrobust,shao2025realworld}. Direct averaging can mix incompatible heart-rate hypotheses, while fixed-weight fusion cannot adapt to changes in candidate reliability for the current video. As shown in Figure~\ref{fig:motivation}(b), such fusion may produce an intermediate estimate unsupported by the candidate hypotheses. Resolving candidate disagreement therefore requires both video-conditioned reliability assessment and consistency checks on the selected candidates before numerical fusion.

To solve this problem, we propose PhysAgent, an inference-time multi-agent framework for candidate verification, as illustrated in Figure~\ref{fig:motivation}(c). A lightweight multimodal large language model (MLLM), e.g., Qwen3-VL-4B \citep{bai2025qwen3vl}, drives agents to integrate visual context, signal quality, and candidate disagreement to assess candidate reliability and guide candidate selection and fusion. We further introduce PCHV, a constrained fusion method that checks execution validity, candidate-cluster support, and harmonic consistency to regulate candidate selection and fusion strategies. It applies correction or fallback when necessary, restricting deterministic numerical fusion to a single hypothesis cluster. The framework operates on outputs from existing base models without retraining them.

Our contributions are threefold. (1) We propose PhysAgent, which treats outputs from different rPPG models as candidate physiological hypotheses and uses agents to assess their reliability through visual and signal evidence, addressing single-model vulnerability and the limitations of naive multi-model fusion. (2) We introduce PCHV, a constrained fusion method that regulates candidate selection and numerical fusion through cluster-support and harmonic-consistency checks, limiting the influence of unsupported agent decisions and extreme candidates on the final estimate. (3) We evaluate PhysAgent on four public rPPG datasets, achieving state-of-the-art overall average performance among the compared methods under intra-dataset and cross-dataset settings.

\section{Related Work}
\textbf{rPPG Estimation and Candidate Fusion.} Early rPPG methods recover pulse-related color variations using blind source separation, chrominance projection, and skin-reflection models \citep{poh2010noncontact,dehaan2013robust,wang2017algorithmic}. Deep methods improve heart-rate and waveform estimation with convolutional attention, spatio-temporal modeling, Transformers, efficient and motion-robust architectures, matrix-factorized attention, and state-space models \citep{chen2018deepphys,yu2019physnet,yu2022physformer,liu2023efficientphys,li2023motionrobust,cho2024factorizephys,zou2025rhythmmamba}, while recent work emphasizes robustness to illumination, domain shift, and complex motion \citep{li2023motionrobust,shao2025realworld}. Classical and deep ensembles can improve robustness when predictors make complementary errors \citep{dietterich2000ensemble,lakshminarayanan2017deep}. In rPPG, however, candidate errors are jointly shaped by training domain, architecture, motion, illumination, and signal quality; mutually inconsistent estimates therefore cannot always be safely averaged.

\vspace{2pt}
\textbf{MLLM Reasoning and Multi-Agent Collaboration.} MLLMs support structured reasoning, task decomposition, tool use, and multimodal interaction. Chain-of-thought prompting and self-consistency improve multi-step reasoning and aggregate diverse reasoning paths \citep{wei2022chain,wang2023selfconsistency}; ReAct and Tree of Thoughts combine reasoning with actions or explicit search \citep{yao2023react,yao2023tree}; and multi-agent debate enables cross-agent deliberation \citep{du2024improving}. Visual instruction tuning and recent MLLMs further extend these capabilities to image-conditioned reasoning \citep{liu2023visual,bai2025qwen3vl}. Most prior methods target discrete decisions, question answering, planning, or semantic visual tasks. In contrast, rPPG is a weak-signal, continuous-valued estimation problem with physiological constraints, making direct MLLM prediction of heart rates or fusion weights difficult to reproduce and constrain physiologically.

\begin{figure*}[!t]
\centering
\includegraphics[width=\textwidth]{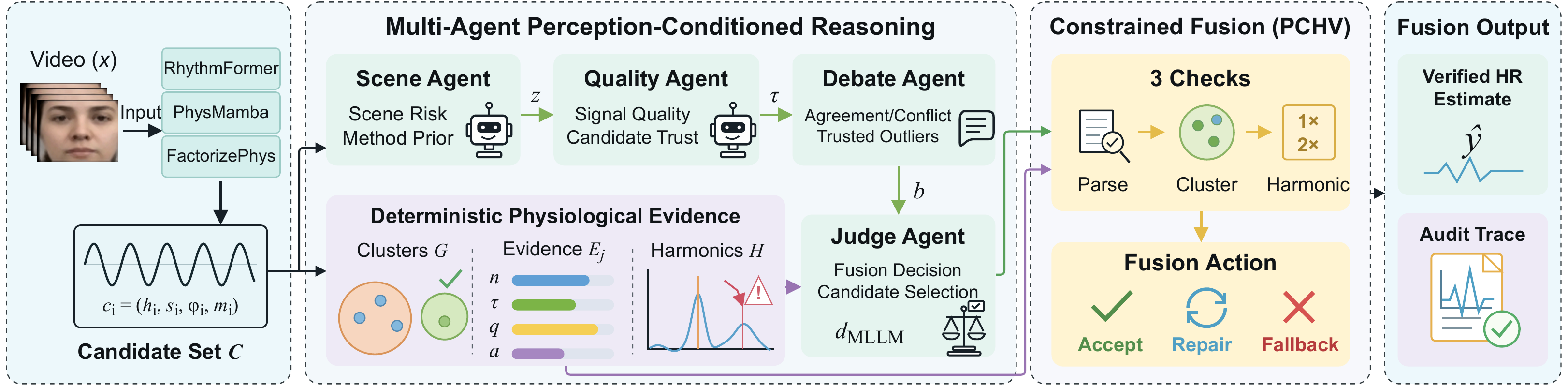}
\caption{Overview of PhysAgent. Base rPPG models produce candidate heart rates and waveforms. Scene, Quality, and Debate agents organize video conditions, signal reliability, and candidate conflicts; Judge combines these assessments with shared cluster and harmonic evidence to select candidates and fusion strategies. PCHV uses the same evidence to constrain deterministic numerical fusion.}
\label{fig:physagent_framework}
\end{figure*}

\section{Method}
The skin-reflection model describes video color variations as jointly affected by pulse, motion, and illumination \citep{wang2017algorithmic}. Inspired by this mechanism, we assess candidate heart rates using both visual context and waveform evidence.

PhysAgent consists of multi-agent candidate reasoning and PCHV-constrained fusion. In the reasoning stage, Scene characterizes video conditions, Quality assesses candidate reliability, Debate examines disagreements, and Judge integrates this information to select candidates and fusion strategies. PCHV then constrains numerical fusion using cluster and harmonic evidence shared with Judge (Figure~\ref{fig:physagent_framework} and Algorithm~\ref{alg:physagent}).

\subsection{Candidate Physiological Hypotheses}
Given a facial video clip \(x\), PhysAgent runs a fixed set of \(K\) base rPPG models, \(\mathcal M=\{m_i\}_{i=1}^{K}\), to obtain heart-rate estimates and associated waveforms and construct the set of candidate physiological hypotheses:
\begin{equation}
\mathcal C=\{c_i\}_{i=1}^{K},\qquad c_i=(h_i,s_i,\phi_i,m_i),
\end{equation}
where \(h_i\) is the estimated heart rate, \(s_i\) is the corresponding rPPG waveform, \(\phi_i\) contains quality features computed from that waveform, and \(m_i\) identifies the source model. Each quality feature is oriented so that a higher value indicates better diagnostic quality:
\begin{equation}
\phi_i=[\widehat{\mathrm{SNR}}_i,P_i,1-H_i,1-M_i,R_i].
\end{equation}
The entries represent normalized SNR, peak prominence, inverse spectral entropy, inverse out-of-band energy, and periodicity, respectively, as defined in Appendix~\ref{app:signal_quality}. The objective is to estimate the final heart rate \(\hat y\) from the video and candidate evidence.

\subsection{Scene-Conditioned Candidate Reasoning}
Periodicity alone is insufficient pulse evidence because motion can also be periodic \citep{wang2017algorithmic}. The Scene Agent therefore examines sampled frames to provide context for interpreting candidate waveforms:
\begin{equation}
z=A_{\mathrm{scene}}(x).
\end{equation}
The scene description includes motion, illumination, skin appearance, occlusion, visual-evidence strength, and model-scene priors; \(v_i\) denotes the assessed suitability of candidate \(c_i\)'s source model. We use these conditions to adjust the relative importance of quality features:
\begin{equation}
\begin{aligned}
\alpha(z)&=\operatorname{softmax}(\log\alpha^0+B\,s(z)),\\
q_i(z)&=\alpha(z)^\top\phi_i .
\end{aligned}
\end{equation}
Here, \(\alpha^0\) is a positive normalized base vector, \(s(z)\) encodes motion, illumination, occlusion, and appearance, and \(B\in\mathbb R^{5\times4}\) maps them to log-weight adjustments. The coefficients are selected experimentally and fixed during inference; Appendix~\ref{app:implementation_details} gives the configuration.

When signal evidence is ambiguous, scene suitability provides supplementary diagnostic information. We scale its influence by visual-evidence strength, candidate disagreement, and the complement of the best signal-quality score:
\begin{equation}
\begin{aligned}
g&=s_{\mathrm{MLLM}}\,d_{\mathrm{cand}}\bigl(1-\max_iq_i(z)\bigr),\\
t_i&=(1-g)q_i(z)+g\,v_i .
\end{aligned}
\end{equation}
Here, \(s_{\mathrm{MLLM}}\) and \(d_{\mathrm{cand}}\) measure visual-evidence strength and candidate disagreement. Thus, scene priors contribute more when visual cues are strong, candidates disagree, and signal quality is weak.

The resulting \(t_i\) accompanies signal features as bounded tie-break evidence, rather than directly setting fusion weights. Quality uses this information to assign candidate trust:
\begin{equation}
\tau_i=A_{\mathrm{qual}}(c_i,z,t_i),\qquad \tau_i\in[0,1].
\end{equation}
Debate then examines the rankings for conflicting modes and overlooked high-trust alternatives, organizing supporting and challenging opinions:
\begin{equation}
b=A_{\mathrm{debate}}(\mathcal C,z,\tau).
\end{equation}
These agents receive candidate and scene evidence; precomputed cluster and harmonic structures are reserved for Judge and PCHV.

\subsection{Hypothesis Evidence and Selection}
Model agreement alone can hide shared errors, while a reliable candidate may remain isolated. We therefore group nearby predictions into competing hypotheses and retain both individual and group support.

Valid candidates are sorted by heart rate and added to the first cluster whose current mean is within \(\epsilon=5\) bpm; the mean is updated after admission, and a new cluster is created if none qualifies. For each \(G_j\in\mathcal G\), we collect \(E_j\):
\begin{equation}
E_j=[n_j,\bar\tau_j,\tau_j^{\max},\bar u_j,u_j^{\max},a_j].
\label{eq:cluster_evidence_vector}
\end{equation}
Here, \(n_j=|G_j|\), bars and maxima summarize members, and \(u_i=\gamma^\top\phi_i\) is scene-independent quality, distinct from \(q_i(z)\). Scene support \(a_j\) measures model-prior support above a neutral reference.

The base cluster support is \(L_j\):
\begin{equation}
\begin{aligned}
L_j={}&\beta_1\tau_j^{\max}+\beta_2\bar\tau_j+\beta_3u_j^{\max}\\
&+\beta_4\bar u_j+\beta_5\rho(n_j)+\beta_6a_j .
\end{aligned}
\label{eq:cluster_support}
\end{equation}
Maximum terms allow a strong candidate to contribute even in a small cluster, while means summarize group support and capped membership limits the influence of model count. The fixed vector \(\beta\) is nonnegative and normalized; Appendix~\ref{app:implementation_details} specifies the coefficients.

When base supports are close, Judge and fallback use contextual support \(\widetilde L_j\) with a bounded diagnostic adjustment; PCHV support and harmonic checks retain \(L_j\). Separated modes can also have near-integer relationships, which we compute before selection:
\begin{equation}
\mathcal H=\{(G_a,G_b):D_{\mathrm{harm}}(G_a,G_b;\mathcal G,\mathcal O)=1\}.
\label{eq:harmonic_detector}
\end{equation}
The detector compares positive centers at enabled harmonic orders \(\mathcal O\) and suppresses pairs dominated in size by a third cluster. Orders, tolerances, and heuristic anchor rules are specified in Appendix~\ref{app:implementation_details}.

Judge combines agent assessments with this shared numerical evidence to determine \(d_{\mathrm{MLLM}}\):
\begin{equation}
d_{\mathrm{MLLM}}=A_{\mathrm{judge}}(\mathcal C,z,\tau,b,\mathcal G,E,\mathcal H)=(\pi,S,w,e),
\end{equation}
where \(\pi\) is the strategy, \(S\) the selected subset, \(w\) the weights, and \(e\) the exclusion set.

\begin{algorithm}[t]
\footnotesize
\caption{PhysAgent inference with constrained fusion.}
\label{alg:physagent}
\begin{algorithmic}[1]
\REQUIRE Video clip \(x\) and base models \(\mathcal{M}\)
\ENSURE Final heart rate \(\hat{y}\) and audit trace \(\mathcal{T}\)
\STATE \(\mathcal{C}\gets\mathrm{CandidateSet}(x,\mathcal{M})\)
\STATE \(z\gets A_{\mathrm{scene}}(x)\)
\STATE \((\tau,b)\gets\mathrm{AgentEvidence}(\mathcal{C},z)\)
\STATE \((\mathcal{G},E,\mathcal{H})\gets
\mathrm{PhysEvidence}(\mathcal{C},z,\tau)\)
\STATE \(d_{\mathrm{MLLM}}\gets
A_{\mathrm{judge}}(\mathcal{C},z,\tau,b,\mathcal{G},E,\mathcal{H})\)
    \IF{Judge is unavailable or its decision is unparsable}
    \STATE \(d^*\gets\mathrm{Fallback}(\mathcal{G},E,\mathcal{H})\)
\ELSE
    \STATE \((v,G^*)\gets
    \mathrm{CheckFusion}(d_{\mathrm{MLLM}},\mathcal{G},E,\mathcal{H})\)
    \STATE \(d^*\gets\mathrm{Normalize}(d_{\mathrm{MLLM}})\) if \(v=\textsc{Accept}\)
    \STATE Otherwise,
    \(d^*\gets\mathrm{Repair}(d_{\mathrm{MLLM}},G^*,\tau)\)
\ENDIF
\STATE \(\hat{y}\gets F(d^*,\mathcal{C})\) using deterministic fusion
\STATE Record evidence and decisions in \(\mathcal{T}\)
\RETURN \(\hat{y},\mathcal{T}\)
\end{algorithmic}
\end{algorithm}
\defcitealias{verkruysse2008remote}{OptExp'08}
\defcitealias{poh2010noncontact}{OptExp'10}
\defcitealias{dehaan2013robust}{T-BME'13}
\defcitealias{wang2017algorithmic}{T-BME'17}
\defcitealias{lee2020metarppg}{ECCV'20}
\defcitealias{yu2019physnet}{BMVC'19}
\defcitealias{zou2025rhythmformer}{PR'25}
\defcitealias{luo2024physmamba}{CCBR'24}
\defcitealias{cho2024factorizephys}{NeurIPS'24}
\defcitealias{zhao2020multiscale}{CVPRW'20}
\defcitealias{song2025spectrogram}{BSPC'25}
\defcitealias{kiddle2023dynamic}{JMIR'23}
\defcitealias{vanputten2023stacked}{CVPRW'23}
\defcitealias{perezgodoy2024desreg}{Neuro'24}
\newcommand{\tabref}[1]{{\scriptsize\citetalias{#1}}}

\begin{table*}[t]
\centering
\footnotesize
\setlength{\tabcolsep}{0.8pt}
\renewcommand{\arraystretch}{1.0}
\captionsetup{font+={color=black}}
\caption{Intra-dataset evaluation (MAE/RMSE in bpm). Best and second-best results are bold and underlined, including ties; rankings use unrounded values where available.}
\color{black}
\label{tab:intra_dataset_results}
\resizebox{0.98\linewidth}{!}{%
\footnotesize
\begin{tabular}{@{}c@{\hspace{4pt}}l*{15}{c}@{}}
\toprule
Paradigm & Method &
\multicolumn{3}{c}{UBFC-rPPG} &
\multicolumn{3}{c}{PURE} &
\multicolumn{3}{c}{BUAA} &
\multicolumn{3}{c}{MMPD} &
\multicolumn{3}{c}{Average} \\
\cmidrule(lr){3-5}\cmidrule(lr){6-8}\cmidrule(lr){9-11}\cmidrule(lr){12-14}\cmidrule(lr){15-17}
& & MAE$\downarrow$ & RMSE$\downarrow$ & \(R\uparrow\)
& MAE$\downarrow$ & RMSE$\downarrow$ & \(R\uparrow\)
& MAE$\downarrow$ & RMSE$\downarrow$ & \(R\uparrow\)
& MAE$\downarrow$ & RMSE$\downarrow$ & \(R\uparrow\)
& MAE$\downarrow$ & RMSE$\downarrow$ & \(R\uparrow\) \\
\midrule
\multirow{10}{*}{\shortstack[c]{Single\\model}} & Green
& 19.73 & 31.00 & 0.37 & 10.09 & 23.85 & 0.34 & 6.89 & 10.39 & 0.60 & 21.68 & 27.69 & -0.01 & 18.12 & 24.59 & 0.33 \\
& ICA
& 16.00 & 25.65 & 0.44 & 4.77 & 16.07 & 0.72 & -- & -- & -- & 18.60 & 24.30 & 0.01 & -- & -- & -- \\
& CHROM
& 4.06 & 8.83 & 0.89 & 5.77 & 14.93 & 0.81 & -- & -- & -- & 13.66 & 18.76 & 0.08 & -- & -- & -- \\
& POS
& 4.08 & 7.72 & 0.92 & 3.67 & 11.82 & 0.88 & -- & -- & -- & 12.36 & 17.71 & 0.18 & -- & -- & -- \\
& Meta-rPPG
& 5.97 & 7.42 & 0.57 & 2.52 & 4.63 & 0.98 & -- & -- & -- & -- & -- & -- & -- & -- & -- \\
& PhysNet
& \underline{0.56} & 2.80 & \underline{0.97} & 0.89 & 4.08 & 0.92 & 14.27 & 18.49 & 0.02 & 9.36 & 16.58 & 0.43 & 8.96 & 14.92 & 0.59 \\
& RhythmFormer
& 0.87 & 3.51 & 0.95 & 1.18 & 4.08 & 0.92 & 10.46 & 14.27 & -0.01 & 5.06 & 11.48 & \underline{0.71} & 5.45 & 10.83 & 0.64 \\
& PhysMamba
& 0.95 & 3.95 & 0.95 & 1.18 & 4.08 & 0.92 & 14.31 & 18.76 & 0.03 & 9.27 & 17.41 & 0.42 & 8.97 & 15.64 & 0.58 \\
& FactorizePhys
& 0.79 & 3.34 & 0.96 & 0.74 & 3.23 & 0.95 & \underline{2.51} & \underline{6.51} & \underline{0.76} & 8.68 & 15.74 & 0.52 & 6.34 & 12.14 & 0.80 \\
& PHASE-Net
& 0.62 & \underline{2.13} & 0.96 & 0.74 & \underline{1.42} & \underline{0.99} & 5.92 & 8.02 & 0.44 & \underline{4.97} & \underline{11.23} & \underline{0.71} & \underline{4.50} & \underline{9.22} & 0.78 \\
\midrule
\multirow{5}{*}{\shortstack[c]{Conventional\\fusion}} & Uniform Mean
& 0.87 & 3.13 & 0.96 & 0.74 & 2.50 & 0.97 & 5.66 & 9.86 & 0.50 & 7.02 & 12.03 & 0.65 & 5.83 & 10.25 & 0.77 \\
& SQI Weighting
& 1.02 & 3.99 & \underline{0.97} & 0.74 & 2.50 & 0.97 & 5.94 & 9.78 & 0.53 & 6.86 & 12.78 & 0.67 & 5.79 & 10.82 & 0.79 \\
& Max-SNR
& 0.95 & 3.96 & 0.95 & \underline{0.59} & 2.89 & 0.96 & 5.23 & 9.73 & 0.50 & 6.26 & 12.59 & 0.64 & 5.25 & 10.70 & 0.76 \\
& Ridge Stacking
& 1.57 & 4.28 & 0.94 & 1.03 & 2.83 & 0.96 & 4.76 & 8.45 & 0.70 & 6.75 & 11.97 & 0.66 & 5.57 & 10.08 & \underline{0.82} \\
& Local Competence
& 1.34 & 4.86 & 0.95 & 0.87 & 3.35 & 0.94 & 5.85 & 9.83 & 0.65 & 6.33 & 12.89 & 0.67 & 5.46 & 11.03 & 0.80 \\
\midrule
Verified fusion & \textbf{PhysAgent (Ours)}
& \textbf{0.51} & \textbf{1.27} & \textbf{0.99}
& \textbf{0.59} & \textbf{0.72} & \textbf{1.00}
& \textbf{2.08} & \textbf{4.77} & \textbf{0.86}
& \textbf{4.73} & \textbf{10.06} & \textbf{0.71}
& \textbf{3.60} & \textbf{7.72} & \textbf{0.89} \\
\bottomrule
\end{tabular}%
}
\end{table*}

\defcitealias{yu2022physformer}{CVPR'22}
\defcitealias{liu2023efficientphys}{WACV'23}

\begin{table*}[t]
\centering
\footnotesize
\setlength{\tabcolsep}{0.8pt}
\renewcommand{\arraystretch}{1.0}
\captionsetup{font+={color=black}}
\caption{Cross-dataset evaluation with three source datasets and one unseen target. Metrics and formatting follow Table~\ref{tab:intra_dataset_results}.}
\color{black}
\label{tab:cross_dataset_leave_one_out}
\resizebox{0.98\linewidth}{!}{%
\footnotesize
\begin{tabular}{@{}c@{\hspace{4pt}}l*{15}{c}@{}}
\toprule
Paradigm & Method &
\multicolumn{3}{c}{Others\(\rightarrow\)U} &
\multicolumn{3}{c}{Others\(\rightarrow\)P} &
\multicolumn{3}{c}{Others\(\rightarrow\)B} &
\multicolumn{3}{c}{Others\(\rightarrow\)M} &
\multicolumn{3}{c}{Average} \\
\cmidrule(lr){3-5}\cmidrule(lr){6-8}\cmidrule(lr){9-11}\cmidrule(lr){12-14}\cmidrule(lr){15-17}
& & MAE$\downarrow$ & RMSE$\downarrow$ & \(R\uparrow\)
& MAE$\downarrow$ & RMSE$\downarrow$ & \(R\uparrow\)
& MAE$\downarrow$ & RMSE$\downarrow$ & \(R\uparrow\)
& MAE$\downarrow$ & RMSE$\downarrow$ & \(R\uparrow\)
& MAE$\downarrow$ & RMSE$\downarrow$ & \(R\uparrow\) \\
\midrule
\multirow{7}{*}{\shortstack[c]{Single\\model}} & Green
& 19.73 & 31.00 & 0.37 & 10.09 & 23.85 & 0.34 & 6.89 & 10.39 & 0.60 & 21.68 & 27.69 & -0.01 & 14.60 & 23.23 & 0.33 \\
& PhysNet
& \underline{2.53} & 7.88 & \underline{0.91} & \underline{7.65} & 17.08 & 0.69 & 4.99 & 9.54 & 0.70 & 12.65 & 19.50 & 0.25 & \underline{6.96} & 13.50 & 0.64 \\
& PhysFormer
& 6.84 & 15.11 & 0.69 & 10.89 & 20.09 & 0.52 & 5.59 & 8.54 & 0.67 & 11.33 & 16.83 & 0.24 & 8.66 & 15.15 & 0.53 \\
& EfficientPhys
& 8.52 & 17.12 & 0.67 & 7.99 & 20.22 & 0.54 & 3.06 & 5.15 & \underline{0.91} & 15.01 & 22.01 & 0.12 & 8.65 & 16.13 & 0.56 \\
& RhythmFormer
& 3.64 & 8.00 & 0.91 & 9.05 & 18.43 & 0.64 & 3.91 & 7.93 & 0.70 & 11.43 & 17.23 & 0.32 & 7.01 & 12.90 & 0.64 \\
& PhysMamba
& 5.98 & \underline{7.64} & 0.90 & 13.39 & 22.33 & 0.60 & 6.26 & 8.16 & 0.71 & 17.10 & 29.70 & 0.26 & 10.68 & 16.96 & 0.62 \\
& PHASE-Net
& 9.67 & 17.66 & 0.59 & 16.48 & 26.78 & 0.11 & \underline{2.75} & 4.71 & 0.91 & 11.62 & 16.84 & 0.26 & 10.13 & 16.50 & 0.47 \\
\midrule
\multirow{5}{*}{\shortstack[c]{Conventional\\fusion}} & Uniform Mean
& 6.18 & 10.37 & 0.86 & 9.26 & \underline{16.90} & \underline{0.70} & 3.06 & \underline{4.56} & 0.90 & \underline{10.98} & \underline{16.38} & \underline{0.32} & 7.37 & \underline{12.05} & \underline{0.69} \\
& SQI Weighting
& 7.11 & 9.69 & 0.87 & 12.55 & 18.55 & 0.64 & 5.89 & 6.99 & 0.78 & 17.41 & 31.64 & 0.20 & 10.74 & 16.72 & 0.62 \\
& Max-SNR
& 5.62 & 8.21 & 0.90 & 13.38 & 21.17 & 0.47 & 5.45 & 6.39 & 0.83 & 18.61 & 33.15 & 0.10 & 10.76 & 17.23 & 0.57 \\
& Ridge Stacking
& 8.42 & 11.81 & 0.81 & 14.01 & 22.42 & 0.51 & 7.54 & 10.69 & 0.60 & 18.33 & 32.48 & 0.17 & 12.08 & 19.35 & 0.52 \\
& Local Competence
& 7.09 & 9.80 & 0.86 & 13.13 & 19.78 & 0.58 & 6.01 & 7.44 & 0.75 & 17.50 & 31.74 & 0.19 & 10.94 & 17.19 & 0.59 \\
\midrule
Verified fusion & \textbf{PhysAgent (Ours)}
& \textbf{1.71} & \textbf{3.09} & \textbf{0.98}
& \textbf{4.90} & \textbf{12.79} & \textbf{0.84}
& \textbf{2.48} & \textbf{3.02} & \textbf{0.98}
& \textbf{10.23} & \textbf{16.13} & \textbf{0.34}
& \textbf{4.83} & \textbf{8.76} & \textbf{0.79} \\
\bottomrule
\end{tabular}%
}
\end{table*}

\begin{table*}[t]
\centering
\footnotesize
\setlength{\tabcolsep}{1.6pt}
\renewcommand{\arraystretch}{1.0}
\captionsetup{font+={color=black}}
\caption{Cross-dataset evaluation on MMPD with two source datasets. Metrics and formatting follow Table~\ref{tab:intra_dataset_results}.}
\color{black}
\label{tab:cross_dataset_limited_mmpd}
\resizebox{0.89\linewidth}{!}{%
\footnotesize
\begin{tabular}{@{}c@{\hspace{4pt}}l*{12}{c}@{}}
\toprule
Paradigm & Method &
\multicolumn{3}{c}{P+B\(\rightarrow\)M} &
\multicolumn{3}{c}{P+U\(\rightarrow\)M} &
\multicolumn{3}{c}{B+U\(\rightarrow\)M} &
\multicolumn{3}{c}{Average} \\
\cmidrule(lr){3-5}\cmidrule(lr){6-8}\cmidrule(lr){9-11}\cmidrule(lr){12-14}
& & MAE$\downarrow$ & RMSE$\downarrow$ & \(R\uparrow\)
& MAE$\downarrow$ & RMSE$\downarrow$ & \(R\uparrow\)
& MAE$\downarrow$ & RMSE$\downarrow$ & \(R\uparrow\)
& MAE$\downarrow$ & RMSE$\downarrow$ & \(R\uparrow\) \\
\midrule
\multirow{7}{*}{\shortstack[c]{Single\\model}} & Green
& 21.68 & 27.69 & -0.01 & 21.68 & 27.69 & -0.01 & 21.68 & 27.69 & -0.01 & 21.68 & 27.69 & -0.01 \\
& PhysNet
& 18.34 & 25.22 & -0.03 & 11.87 & 18.77 & \underline{0.27} & \underline{14.23} & 20.18 & 0.07 & 14.81 & 21.39 & 0.10 \\
& PhysFormer
& 22.28 & 35.94 & 0.00 & 18.97 & 32.42 & 0.06 & 19.94 & 33.51 & 0.00 & 20.40 & 33.96 & 0.02 \\
& EfficientPhys
& 23.91 & 38.03 & 0.01 & 21.42 & 36.16 & 0.03 & 21.41 & 35.79 & 0.08 & 22.25 & 36.66 & 0.04 \\
& RhythmFormer
& 18.01 & 32.10 & 0.19 & 17.29 & 29.74 & 0.24 & 17.02 & 30.33 & 0.17 & 17.44 & 30.72 & 0.20 \\
& PhysMamba
& 17.71 & 31.20 & 0.15 & 16.81 & 29.49 & \underline{0.27} & 17.22 & 29.74 & \underline{0.24} & 17.25 & 30.15 & \underline{0.22} \\
& PHASE-Net
& \underline{11.94} & \underline{17.45} & \underline{0.21} & \underline{11.56} & \underline{17.54} & 0.20 & 14.90 & \underline{19.36} & 0.14 & \underline{12.80} & \underline{18.12} & 0.18 \\
\midrule
\multirow{5}{*}{\shortstack[c]{Conventional\\fusion}} & Uniform Mean
& 20.17 & 34.29 & 0.09 & 17.30 & 31.01 & 0.16 & 17.65 & 31.75 & 0.12 & 18.37 & 32.35 & 0.12 \\
& SQI Weighting
& 20.02 & 34.22 & 0.09 & 17.19 & 30.90 & 0.17 & 17.44 & 31.58 & 0.13 & 18.22 & 32.23 & 0.13 \\
& Max-SNR
& 20.16 & 34.93 & 0.06 & 18.30 & 32.00 & 0.16 & 17.82 & 31.68 & 0.15 & 18.76 & 32.87 & 0.12 \\
& Ridge Stacking
& 18.16 & 32.32 & 0.19 & 23.32 & 38.35 & -0.02 & 19.45 & 33.32 & 0.02 & 20.31 & 34.66 & 0.06 \\
& Local Competence
& 19.79 & 33.90 & 0.12 & 17.36 & 31.13 & 0.16 & 18.40 & 32.59 & 0.06 & 18.51 & 32.54 & 0.11 \\
\midrule
Verified fusion & \textbf{PhysAgent (Ours)}
& \textbf{11.57} & \textbf{17.10} & \textbf{0.27}
& \textbf{10.98} & \textbf{16.67} & \textbf{0.33}
& \textbf{10.43} & \textbf{16.16} & \textbf{0.42}
& \textbf{10.99} & \textbf{16.64} & \textbf{0.34} \\
\bottomrule
\end{tabular}%
}
\end{table*}

\begin{table*}[t]
\centering
\footnotesize
\setlength{\tabcolsep}{1.6pt}
\renewcommand{\arraystretch}{1.0}
\captionsetup{font+={color=black}}
\caption{Cross-dataset evaluation on BUAA-MIHR with two source datasets. Metrics and formatting follow Table~\ref{tab:intra_dataset_results}.}
\color{black}
\label{tab:cross_dataset_limited_buaa}
\resizebox{0.87\linewidth}{!}{%
\footnotesize
\begin{tabular}{@{}c@{\hspace{4pt}}l*{12}{c}@{}}
\toprule
Paradigm & Method &
\multicolumn{3}{c}{P+M\(\rightarrow\)B} &
\multicolumn{3}{c}{M+U\(\rightarrow\)B} &
\multicolumn{3}{c}{P+U\(\rightarrow\)B} &
\multicolumn{3}{c}{Average} \\
\cmidrule(lr){3-5}\cmidrule(lr){6-8}\cmidrule(lr){9-11}\cmidrule(lr){12-14}
& & MAE$\downarrow$ & RMSE$\downarrow$ & \(R\uparrow\)
& MAE$\downarrow$ & RMSE$\downarrow$ & \(R\uparrow\)
& MAE$\downarrow$ & RMSE$\downarrow$ & \(R\uparrow\)
& MAE$\downarrow$ & RMSE$\downarrow$ & \(R\uparrow\) \\
\midrule
\multirow{7}{*}{\shortstack[c]{Single\\model}} & Green
& 6.89 & 10.39 & 0.60 & 6.89 & 10.39 & 0.60 & 6.89 & 10.39 & 0.60 & 6.89 & 10.39 & 0.60 \\
& PhysNet
& \underline{4.32} & 8.38 & 0.75 & 7.76 & 14.16 & 0.44 & 7.51 & 11.79 & 0.45 & 6.53 & 11.44 & 0.55 \\
& PhysFormer
& 5.83 & 7.67 & 0.76 & 8.50 & 12.82 & 0.30 & 5.21 & 6.07 & 0.84 & 6.51 & 8.85 & 0.64 \\
& EfficientPhys
& 7.05 & 9.23 & 0.73 & 5.43 & 7.91 & 0.77 & 5.16 & 6.06 & 0.86 & 5.88 & 7.73 & 0.78 \\
& RhythmFormer
& 9.32 & 14.04 & 0.08 & 7.92 & 11.34 & 0.44 & 9.79 & 13.80 & 0.37 & 9.01 & 13.06 & 0.30 \\
& PhysMamba
& 7.35 & 9.56 & 0.62 & 8.59 & 13.44 & 0.41 & 7.97 & 11.65 & 0.51 & 7.97 & 11.55 & 0.51 \\
& PHASE-Net
& 9.78 & 15.53 & 0.14 & \underline{3.01} & \underline{5.13} & 0.78 & 7.79 & 12.51 & 0.55 & 6.86 & 11.06 & 0.49 \\
\midrule
\multirow{5}{*}{\shortstack[c]{Conventional\\fusion}} & Uniform Mean
& 6.11 & 7.08 & 0.79 & 6.39 & 7.81 & 0.72 & 6.08 & 6.91 & 0.79 & 6.19 & 7.27 & 0.77 \\
& SQI Weighting
& 5.93 & \underline{6.85} & \underline{0.80} & 6.26 & 7.54 & 0.75 & 5.95 & 6.72 & 0.80 & 6.05 & 7.04 & 0.78 \\
& Max-SNR
& 6.10 & 7.74 & 0.75 & 5.74 & 7.21 & \underline{0.78} & \underline{4.72} & \underline{5.53} & \underline{0.87} & \underline{5.52} & \underline{6.83} & \underline{0.80} \\
& Ridge Stacking
& 8.84 & 12.40 & 0.45 & 9.15 & 13.42 & 0.44 & 5.68 & 7.36 & 0.83 & 7.89 & 11.06 & 0.57 \\
& Local Competence
& 6.26 & 7.80 & 0.72 & 6.53 & 8.19 & 0.69 & 6.02 & 6.83 & 0.80 & 6.27 & 7.60 & 0.74 \\
\midrule
Verified fusion & \textbf{PhysAgent (Ours)}
& \textbf{3.15} & \textbf{4.89} & \textbf{0.91}
& \textbf{2.88} & \textbf{4.25} & \textbf{0.92}
& \textbf{2.69} & \textbf{3.33} & \textbf{0.96}
& \textbf{2.91} & \textbf{4.15} & \textbf{0.93} \\
\bottomrule
\end{tabular}%
}
\end{table*}

\subsection{PCHV-Constrained Fusion}
A plausible candidate selection can still mix different heart-rate modes, placing a fused estimate between incompatible hypotheses. PCHV therefore checks executability, single-cluster support, and harmonic-anchor consistency using shared evidence:
\begin{equation}
d^*=V(d_{\mathrm{MLLM}},\mathcal G,E,\mathcal H).
\end{equation}
With valid nonempty cluster evidence, an empty or cross-cluster selection is initially repaired to the best base-support cluster \(G_{\mathrm{base}}\):
\begin{equation}
G_{\mathrm{base}}=\operatorname*{arg\,max}^{\mathrm{lex}}_{G_j\in\mathcal G}(L_j,\tau_j^{\max},n_j).
\end{equation}
Remaining ties retain construction order. A sufficiently large support gap triggers replacement, while an applicable harmonic-anchor check can override the repair target.

If Judge is unavailable or its decision is unparsable, fallback ranks clusters using \(\widetilde L_j\). Appendix~\ref{app:implementation_details} specifies thresholds and check order.

Let \(G^*\) denote the repair or fallback target:
\begin{equation}
d^*=\begin{cases}
\mathrm{Normalize}(d_{\mathrm{MLLM}}), & \text{\small checks passed,}\\
\mathrm{Repair}(d_{\mathrm{MLLM}},G^*), & \begin{gathered}\text{\small parseable decision}\\\text{\small requiring correction,}\end{gathered}\\
\mathrm{Fallback}(G^*), & \begin{gathered}\text{\small Judge unavailable or}\\\text{\small decision unparsable.}\end{gathered}
\end{cases}
\end{equation}
Repair and fallback restrict the selection to \(G^*\):
\begin{equation}
S^*=G^*,\qquad e^*=\mathcal C\setminus G^*.
\end{equation}
For weighted fusion, acceptance normalizes surviving positive Judge weights, repair uses candidate trust, and fallback additionally incorporates raw quality:
\begin{equation}
\begin{aligned}
w_i^*&=\frac{r_i}{\sum_{c_k\in S^*}r_k},\\
r_i&=\begin{cases}
w_i,&\text{accept},\\
\max(\tau_i,\eta),&\text{repair},\\
\max(\tau_i,u_i,\eta),&\text{fallback}.
\end{cases}
\end{aligned}
\end{equation}
The positive floor \(\eta\) prevents zero total weight; a singleton returns its heart rate directly. The final estimate is computed deterministically by \(F\):
\begin{equation}
\hat y=F(d^*,\mathcal C).
\end{equation}
\(F\) supports single selection, weighted mean, median, and trimmed mean within the selected cluster. With valid cluster evidence and finite normalized nonnegative weights, the output remains within the selected cluster's candidate range; exceptional-input handling is specified in Appendix~\ref{app:implementation_details}.

\section{Experiments}
We evaluate PhysAgent on four standard rPPG datasets: UBFC-rPPG \citep{bobbia2019unsupervised}, PURE \citep{stricker2014noncontact}, BUAA-MIHR \citep{xi2020image}, and MMPD \citep{tang2023mmpd}. Intra-dataset evaluation uses conventional within-dataset splits, whereas cross-dataset evaluation assesses generalization to unseen target datasets using base models trained only on source data. We report mean absolute error (MAE) and root mean squared error (RMSE) in bpm (lower is better), and Pearson's correlation coefficient \(R\) (higher is better). Appendix~\ref{app:datasets_protocols} details the datasets, base-model sources, and evaluation protocols.

\paragraph{Comparison Methods.}
\textit{Base models.}
We include four classical rPPG methods: Green \citep{verkruysse2008remote}, ICA \citep{poh2010noncontact}, CHROM \citep{dehaan2013robust}, and POS \citep{wang2017algorithmic}; and \reviewchange{eight learning-based models}: PhysNet \citep{yu2019physnet}, Meta-rPPG \citep{lee2020metarppg}, PhysFormer \citep{yu2022physformer}, EfficientPhys \citep{liu2023efficientphys}, PhysMamba \citep{luo2024physmamba}, FactorizePhys \citep{cho2024factorizephys}, RhythmFormer \citep{zou2025rhythmformer}, and PHASE-Net \citep{zhao2026phasenet}.
\textit{Fusion methods.}
We compare five conventional fusion baselines adapted from established aggregation principles: Uniform Mean \citep{zhao2020multiscale}, Max-SNR \citep{kiddle2023dynamic}, Ridge Stacking \citep{vanputten2023stacked}, Local Competence \citep{perezgodoy2024desreg}, and SQI Weighting \citep{song2025spectrogram}.

\vspace{2pt}
PhysAgent organizes candidate disagreement into HR hypotheses, combining
video context, signal quality, and candidate relationships for hypothesis
selection and within-cluster fusion.

\subsection{Intra-Dataset Evaluation}

Table~\ref{tab:intra_dataset_results} evaluates \reviewchange{single models}, five conventional fusion baselines, and PhysAgent under common intra-dataset splits; PHASE-Net is our re-evaluation and is not part of the aligned candidate pool. Every multi-model method uses the same RhythmFormer, PhysMamba, and FactorizePhys candidates, which PhysAgent does not retrain. We report MAE, RMSE, and Pearson \(R\); average MAE/RMSE is sample-weighted, while average \(R\) is the mean of the four dataset-level correlations. Appendix~\ref{app:conventional_fusion} provides full rules. PhysAgent achieves the best average MAE/RMSE/\(R\) of 3.60/7.72/0.89 and, using unrounded values, ranks first in all three metrics on every dataset. It also outperforms our PHASE-Net re-evaluation on all twelve dataset-level metrics.
The consistent \reviewchange{intra-dataset} gains indicate that reasoning about candidate disagreement for each sample helps translate complementary information from \reviewchange{multiple models} into more accurate HR predictions.

\begin{figure}[!t]
\centering
\includegraphics[width=\linewidth]{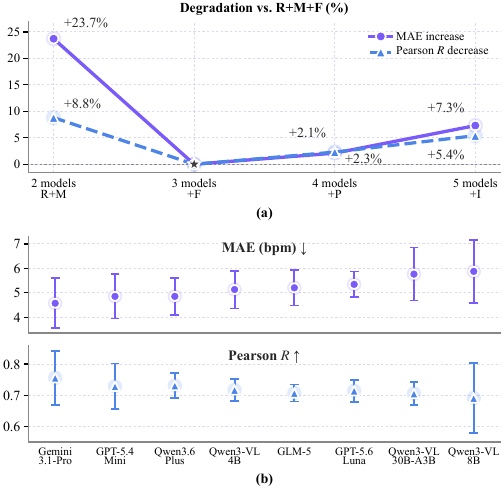}
\captionsetup{font+={color=black}}
\caption{Sensitivity analysis. (a) Candidate-pool expansion relative to the Table~\ref{tab:intra_dataset_results} \(R+M+F\) reference. (b) Sampled MLLM-backbone robustness in MAE and Pearson \(R\); markers and error bars denote the mean and standard deviation over five fixed 40-sample batches.}
\label{fig:sensitivity_analysis}
\end{figure}

\subsection{Cross-Dataset Evaluation}

We further evaluate cross-dataset generalization under the leave-one-out protocol, where one dataset is held out as the unseen target and the remaining datasets serve as source domains. Table~\ref{tab:cross_dataset_leave_one_out} compares \reviewchange{single models}, including our PHASE-Net re-evaluation, five conventional fusion baselines, and PhysAgent. The conventional fusion methods aggregate source-trained PhysFormer, EfficientPhys, and RhythmFormer outputs, with learned components fitted only on source-domain train/validation data. These fusion rules cover equal-weight aggregation, quality-driven selection or weighting, and learned global weights or local competence. Their performance remains target-dependent and does not consistently improve over the strongest \reviewchange{single model}. PhysAgent achieves the best average MAE/RMSE/\(R\) of 4.83/8.76/0.79 and, using unrounded values, ranks first in all three metrics on every target dataset. It also outperforms PHASE-Net on all twelve target-level metrics, supporting stable candidate verification across unseen domains.

Tables~\ref{tab:cross_dataset_limited_mmpd} and~\ref{tab:cross_dataset_limited_buaa} further evaluate the limited-source setting, where only two source domains are available for transfer to a specified target domain. Both tables compare the five conventional fusion baselines and our PHASE-Net re-evaluation under the same source-only constraints. On MMPD, PhysAgent achieves the best average MAE/RMSE/\(R\) of 10.99/16.64/0.34 and ranks first in all three metrics for every source-domain combination. On BUAA-MIHR, PhysAgent achieves the best average MAE/RMSE/\(R\) of 2.91/4.15/0.93 and, using unrounded values, likewise ranks first in all three metrics for every source-domain combination. It also outperforms PHASE-Net on all nine protocol-level metrics. These results show that PhysAgent remains robust under limited-source shifts and achieves the best average metrics on both targets.

\subsection{Ablation Study}

Table~\ref{tab:ablation_results} reports component ablations relative to the Full references in Tables~\ref{tab:intra_dataset_results} and~\ref{tab:cross_dataset_leave_one_out}.

\noindent\textbf{Agent Collaboration.} The Scene, Quality, Debate, and MLLM-Judge ablation configurations have higher reported errors under both intra-dataset and cross-dataset evaluation. The MLLM-Judge replacement shows the largest intra-dataset MAE/RMSE increases (62.2\%/63.7\%), whereas removing Debate shows the largest cross-dataset increases (70.0\%/53.3\%), highlighting the roles of candidate-reliability assessment and conflict analysis across the two settings.

\noindent\textbf{PCHV-Constrained Fusion.} Removing PCHV and directly executing the Judge's fusion decisions increases intra-dataset MAE/RMSE by 42.5\%/39.1\% and cross-dataset MAE/RMSE by 55.4\%/48.8\%. These results support checking candidate consistency and correcting constraint-violating fusion decisions before numerical execution.

\noindent\textbf{Rules-Only Control.} The configuration without MLLM agents retains deterministic clustering, rule-based cluster selection, verification, and within-cluster fusion, but has higher reported errors than the Full reference. Together with the component ablations, this indicates that the tested fixed rules do not fully substitute for the agents' joint assessment of scene context, signal quality, and candidate disagreement.

Appendices~\ref{app:ablation_replacements} and~\ref{app:empirical_reliability} detail ablation settings and reliability analyses, respectively.

\begin{table}[t]
\centering
\scriptsize
\setlength{\tabcolsep}{1pt}
\renewcommand{\arraystretch}{1.00}

\captionsetup{font+={color=black}}
\caption{Component ablations and a rules-only control (MAE/RMSE in bpm). Parentheses show arithmetic changes relative to the reported Full averages from Tables~\ref{tab:intra_dataset_results} and~\ref{tab:cross_dataset_leave_one_out}. Rules-only results use archived candidates; Appendix~\ref{app:ablation_replacements} details configurations and aggregation.}
\label{tab:ablation_results}
\resizebox{\linewidth}{!}{%
\scriptsize
\begin{tabular}{@{}lcccc@{}}
\toprule
& \multicolumn{2}{c}{\reviewchange{Intra-dataset}} & \multicolumn{2}{c}{\reviewchange{Cross-dataset}} \\
\cmidrule(lr){2-3}\cmidrule(lr){4-5}
Variant & MAE\(\downarrow\) & RMSE\(\downarrow\) & MAE\(\downarrow\) & RMSE\(\downarrow\) \\
\midrule
\textbf{Full PhysAgent} & \textbf{3.60} & \textbf{7.72} & \textbf{4.83} & \textbf{8.76} \\
w/o Scene Agent & 4.67 (+29.7\%) & 10.71 (+38.7\%) & 7.79 (+61.3\%) & 13.35 (+52.5\%) \\
w/o Quality Agent & 4.78 (+32.8\%) & 10.42 (+35.0\%) & 7.18 (+48.6\%) & 11.95 (+36.5\%) \\
w/o Debate Agent & 4.91 (+36.4\%) & 10.86 (+40.7\%) & 8.21 (+70.0\%) & 13.42 (+53.3\%) \\
w/o MLLM Judge & 5.84 (+62.2\%) & 12.64 (+63.7\%) & 6.56 (+35.9\%) & 12.06 (+37.7\%) \\
w/o PCHV & 5.13 (+42.5\%) & 10.74 (+39.1\%) & 7.51 (+55.4\%) & 13.03 (+48.8\%) \\
w/o all MLLM agents & 5.30 \reviewchange{(+47.2\%)} & 11.15 \reviewchange{(+44.4\%)} & 7.32 \reviewchange{(+51.5\%)} & 12.53 \reviewchange{(+43.0\%)} \\
\bottomrule
\end{tabular}%
}
\end{table}

\begin{figure*}[!t]
\centering
\includegraphics[width=\textwidth]{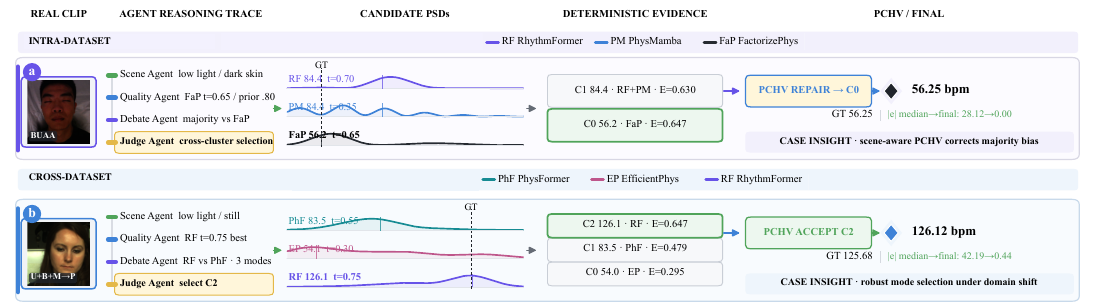}
\captionsetup{font+={color=black},skip=3pt}
\caption{Representative intra-dataset and cross-dataset cases. (a) PCHV corrects a cross-cluster selection to the evidence-supported minority hypothesis. (b) The agents select RhythmFormer's prediction, and PCHV accepts it. Dashed lines indicate reference heart rates (GT). U, B, M, and P denote UBFC-rPPG, BUAA-MIHR, MMPD, and PURE, respectively.}
\label{fig:case_analysis}
\end{figure*}

\subsection{Further Analysis}

\begin{table}[t]
\centering
\scriptsize
\setlength{\tabcolsep}{2.2pt}
\renewcommand{\arraystretch}{1.0}
\captionsetup{font+={color=black}}
\caption{MMPD attribute results (MAE/RMSE, bpm; \(\downarrow\)).}
\color{black}
\label{tab:mmpd_attribute_robustness}
\resizebox{\linewidth}{!}{%
\scriptsize
\begin{tabular}{@{}lcccc@{}}
\multicolumn{5}{c}{(a) Skin Tones} \\
\midrule
Method & T3 & T4 & T5 & T6 \\
\midrule
PhysFormer
& 6.11/11.21
& 5.92/10.00
& 5.87/13.12
& 6.81/11.21 \\
RhythmFormer
& 5.12/11.23
& 5.46/9.14
& 6.32/11.87
& 6.26/9.99 \\
\textbf{PhysAgent (Ours)}
& \textbf{1.64}/\textbf{4.34}
& \textbf{4.10}/\textbf{6.51}
& \textbf{4.55}/\textbf{9.80}
& \textbf{4.80}/\textbf{7.40} \\
\midrule
\multicolumn{5}{c}{(b) Lighting Conditions} \\
\midrule
Method & LED-L & LED-H & Incand. & Natural \\
\midrule
PhysFormer
& 6.36/11.72
& 5.12/9.39
& 5.71/12.73
& 6.61/11.36 \\
RhythmFormer
& 5.85/11.71
& 4.46/8.98
& 3.64/11.87
& 5.65/12.31 \\
\textbf{PhysAgent (Ours)}
& \textbf{4.28}/\textbf{8.95}
& \textbf{3.45}/\textbf{7.55}
& \textbf{3.30}/\textbf{9.95}
& \textbf{4.11}/\textbf{8.03} \\
\bottomrule
\end{tabular}%
}
\end{table}

\paragraph{\reviewchange{Attribute-wise Robustness.}}
\reviewchange{We group MMPD by four skin-tone types and four lighting conditions, and PhysAgent achieves lower MAE and RMSE than PhysFormer and RhythmFormer in all eight groups (Table~\ref{tab:mmpd_attribute_robustness}). Its MAE/RMSE is 3.30/9.95 bpm under incandescent lighting and 4.11/8.03 bpm under natural lighting, demonstrating adaptability to appearance and illumination variations. Compared with RhythmFormer, PhysAgent reduces MAE by approximately 28.0\% and 23.3\% in the T5 and T6 groups, respectively. These gains extend across skin-tone conditions, with improvements in both MAE and RMSE.}

\vspace{-7pt}
\paragraph{\reviewchange{Scaling Analysis.}}
\reviewchange{With Qwen3-VL-4B and all other settings fixed, we first add FactorizePhys to the RhythmFormer and PhysMamba pool, then add PhysNet and iBVPNet \citep{joshi2024ibvp} (Figure~\ref{fig:sensitivity_analysis}(a)). Relative to the three-model pool, MAE increases by 23.7\%, 2.1\%, and 7.3\% for the two-, four-, and five-model pools, respectively, with Pearson \(R\) also decreasing. Expanding the candidate pool changes both the available heart-rate predictions and inter-model agreement, so the value of an additional model depends on whether it compensates for failures of existing models and provides complementary predictions.}

\reviewchange{We fix the three-model candidate pool and proportionally sample 200 examples from the four datasets, dividing them into five non-overlapping batches of 40, to compare MLLM backbones (Figure~\ref{fig:sensitivity_analysis}(b)). Qwen3-VL-4B achieves lower mean MAE than the 8B and 30B-A3B variants from the same series, but higher mean MAE than backbones such as Gemini-3.1-Pro.}

\reviewchange{Balancing estimation performance and local inference cost, we select Qwen3-VL-4B as our default backbone. Its agent stage averages 18.45K tokens and 12.59 seconds of serial processing per sample; Appendix~\ref{app:token_cost} details the cost measurements and estimates. The lower mean MAE obtained by stronger backbones in this comparison also highlights the potential to improve PhysAgent's heart-rate estimation by using a more capable MLLM.}

\vspace{-7pt}
\paragraph{\reviewchange{Reliability and Output Stability.}}
\reviewchange{On a fixed, dataset-balanced cohort of 479 samples, we repeat the complete pipeline with three random seeds, obtaining Judge cluster agreement of 82.3\%, final-HR agreement of 96.5\%, and a mean sample-level HR standard deviation of 0.34 bpm. This indicates that final heart-rate outputs are more stable than intermediate cluster-selection decisions.}

\reviewchange{In a separate controlled analysis of 1,803 samples, the mean absolute errors are 1.19, 6.66, and 11.81 bpm for samples with one, two, and three candidate clusters, respectively. The increasing error with cluster count reflects the association between candidate conflict and estimation difficulty. With upstream agent decisions fixed, PCHV reduces MAE in the controlled replay from 4.74 to 4.55 bpm, demonstrating the value of checking and correcting fusion decisions before numerical execution. In the same replay, 305 of the 307 repairs occur in multi-cluster samples, while absolute error remains unchanged for 84.9\% of the cohort. These results link PCHV intervention to candidate disagreement and its associated estimation difficulty. Appendix~\ref{app:empirical_reliability} provides the complete statistics and protocols.}

\vspace{-10pt}
\paragraph{\reviewchange{Case Analysis.}}
\reviewchange{Figure~\ref{fig:case_analysis}(a) shows an intra-dataset case where RhythmFormer and PhysMamba agree at 84.38 bpm, while FactorizePhys gives 56.25 bpm. Scene-conditioned and signal-quality evidence supports the minority hypothesis, and PCHV corrects the cross-cluster selection to yield 56.25 bpm, matching the reference heart rate. Here, the two agreeing predictions deviate from the reference, showing the importance of assessing model agreement using evidence from the current video. Joint assessment of scene and signal quality allows the reliable minority prediction to be retained.}

\reviewchange{Figure~\ref{fig:case_analysis}(b) shows a cross-dataset case where the three models give different heart-rate predictions, and the agents select RhythmFormer's prediction of 126.12 bpm. PCHV accepts this selection, and the final result differs from the reference heart rate of 125.68 bpm by 0.44 bpm. The large disagreement among the three models means that direct averaging would mix inconsistent predictions and produce an intermediate heart rate unsupported by the candidates. This case illustrates the agents' role in resolving candidate disagreement, while PCHV retains their constraint-compliant selection for numerical execution.}

\reviewchange{Additional cases are provided in Appendix~\ref{app:case_studies}. Limitations are discussed in Appendix~\ref{app:limitations}.}

\FloatBarrier

\section{Conclusion}
We presented PhysAgent, a multi-agent framework for reliable remote heart-rate estimation. PhysAgent uses multiple agents to generate structured fusion decisions and applies deterministic physiological constraints for verification and numerical execution, improving auditability and reducing unsupported cross-cluster fusion. Experiments and ablations show that PhysAgent achieves the best overall average performance in both intra- and cross-dataset settings and confirm the complementary roles of multi-agent reasoning and deterministic verification. 

\section*{LLM Usage Statement}
PhysAgent uses Qwen3-VL-4B for the structured scene, quality, debate, and Judge agents described in this paper. The backbone-sensitivity study additionally evaluates Gemini 3.1-Pro, GPT-5.4 Mini, Qwen3.6 Plus, GLM-5, GPT-5.6 Luna, Qwen3-VL-30B-A3B, and Qwen3-VL-8B. OpenAI Codex assisted with the CVPR template migration, LaTeX formatting, English editing of the ablation, reliability, and limitations discussions, layout checks, and preparation of this usage statement. The authors are responsible for the content and scientific claims.

{\small
\bibliographystyle{ieeenat_fullname}
\bibliography{physagent}

@article{poh2010noncontact,
  author = {Poh, Ming-Zher and McDuff, Daniel J. and Picard, Rosalind W.},
  title = {Non-contact, Automated Cardiac Pulse Measurements Using Video Imaging and Blind Source Separation},
  journal = {Optics Express},
  volume = {18},
  number = {10},
  pages = {10762--10774},
  year = {2010},
  doi = {10.1364/OE.18.010762}
}

@article{verkruysse2008remote,
  author = {Verkruysse, Wim and Svaasand, Lars O. and Nelson, J. Stuart},
  title = {Remote Plethysmographic Imaging Using Ambient Light},
  journal = {Optics Express},
  volume = {16},
  number = {26},
  pages = {21434--21445},
  year = {2008},
  doi = {10.1364/OE.16.021434}
}

@article{dehaan2013robust,
  author = {de Haan, Gerard and Jeanne, Vincent},
  title = {Robust Pulse Rate From Chrominance-Based rPPG},
  journal = {IEEE Transactions on Biomedical Engineering},
  volume = {60},
  number = {10},
  pages = {2878--2886},
  year = {2013},
  doi = {10.1109/TBME.2013.2266196}
}

@article{wang2017algorithmic,
  author = {Wang, Wenjin and den Brinker, Albertus C. and Stuijk, Sander and de Haan, Gerard},
  title = {Algorithmic Principles of Remote PPG},
  journal = {IEEE Transactions on Biomedical Engineering},
  volume = {64},
  number = {7},
  pages = {1479--1491},
  year = {2017},
  doi = {10.1109/TBME.2016.2609282}
}

@article{bobbia2019unsupervised,
  author = {Bobbia, Serge and Macwan, Richard and Benezeth, Yannick and Mansouri, Alamin and Dubois, Julien},
  title = {Unsupervised Skin Tissue Segmentation for Remote Photoplethysmography},
  journal = {Pattern Recognition Letters},
  volume = {124},
  pages = {82--90},
  year = {2019},
  doi = {10.1016/j.patrec.2017.10.017}
}

@inproceedings{stricker2014noncontact,
  author = {Stricker, Ronny and Mueller, Steffen and Gross, Horst-Michael},
  title = {Non-Contact Video-Based Pulse Rate Measurement on a Mobile Service Robot},
  booktitle = {Proceedings of the 23rd IEEE International Symposium on Robot and Human Interactive Communication},
  pages = {1056--1062},
  year = {2014},
  doi = {10.1109/ROMAN.2014.6926392}
}

@inproceedings{xi2020image,
  author = {Xi, Lin and Chen, Weihai and Zhao, Changchen and Wu, Xingming and Wang, Jianhua},
  title = {Image Enhancement for Remote Photoplethysmography in a Low-Light Environment},
  booktitle = {Proceedings of the 15th IEEE International Conference on Automatic Face and Gesture Recognition},
  pages = {1--7},
  year = {2020},
  doi = {10.1109/FG47880.2020.00076}
}

@inproceedings{tang2023mmpd,
  author = {Tang, Jiankai and Chen, Kequan and Wang, Yuntao and Shi, Yuanchun and Patel, Shwetak and McDuff, Daniel and Liu, Xin},
  title = {{MMPD}: Multi-Domain Mobile Video Physiology Dataset},
  booktitle = {Proceedings of the 45th Annual International Conference of the IEEE Engineering in Medicine and Biology Society},
  pages = {1--5},
  year = {2023},
  doi = {10.1109/EMBC40787.2023.10340857}
}

@inproceedings{chen2018deepphys,
  author = {Chen, Weixuan and McDuff, Daniel},
  title = {DeepPhys: Video-Based Physiological Measurement Using Convolutional Attention Networks},
  booktitle = {Proceedings of the European Conference on Computer Vision (ECCV)},
  pages = {349--365},
  year = {2018}
}

@inproceedings{yu2019physnet,
  author = {Yu, Zitong and Li, Xiaobai and Zhao, Guoying},
  title = {Remote Photoplethysmograph Signal Measurement from Facial Videos Using Spatio-Temporal Networks},
  booktitle = {Proceedings of the British Machine Vision Conference (BMVC)},
  pages = {277},
  year = {2019}
}

@inproceedings{lee2020metarppg,
  author = {Lee, Eugene and Chen, Evan and Lee, Chen-Yi},
  title = {{Meta-rPPG}: Remote Heart Rate Estimation Using a Transductive Meta-Learner},
  booktitle = {Proceedings of the European Conference on Computer Vision (ECCV)},
  pages = {392--409},
  year = {2020},
  doi = {10.1007/978-3-030-58583-9_24}
}

@article{zou2025rhythmformer,
  author = {Zou, Bochao and Guo, Zizheng and Chen, Jiansheng and Zhuo, Junbao and Huang, Weiran and Ma, Huimin},
  title = {{RhythmFormer}: Extracting Patterned {rPPG} Signals Based on Periodic Sparse Attention},
  journal = {Pattern Recognition},
  volume = {164},
  pages = {111511},
  year = {2025},
  doi = {10.1016/j.patcog.2025.111511}
}

@inproceedings{luo2024physmamba,
  author = {Luo, Chaoqi and Xie, Yiping and Yu, Zitong},
  title = {{PhysMamba}: Efficient Remote Physiological Measurement with {SlowFast} Temporal Difference {Mamba}},
  booktitle = {Biometric Recognition (CCBR 2024)},
  series = {Lecture Notes in Computer Science},
  volume = {15353},
  pages = {248--259},
  year = {2025},
  doi = {10.1007/978-981-96-1071-6_23}
}

@inproceedings{yu2022physformer,
  author = {Yu, Zitong and Shen, Yuming and Shi, Jingang and Zhao, Hengshuang and Torr, Philip H. S. and Zhao, Guoying},
  title = {PhysFormer: Facial Video-Based Physiological Measurement With Temporal Difference Transformer},
  booktitle = {Proceedings of the IEEE/CVF Conference on Computer Vision and Pattern Recognition (CVPR)},
  pages = {4186--4196},
  year = {2022}
}

@inproceedings{liu2023efficientphys,
  author = {Liu, Xin and Hill, Brian and Jiang, Ziheng and Patel, Shwetak and McDuff, Daniel},
  title = {EfficientPhys: Enabling Simple, Fast and Accurate Camera-Based Cardiac Measurement},
  booktitle = {Proceedings of the IEEE/CVF Winter Conference on Applications of Computer Vision (WACV)},
  pages = {5008--5017},
  year = {2023}
}

@article{li2023motionrobust,
  author = {Li, Jianwei and Yu, Zitong and Shi, Jingang},
  title = {Learning Motion-Robust Remote Photoplethysmography Through Arbitrary Resolution Videos},
  journal = {Proceedings of the AAAI Conference on Artificial Intelligence},
  volume = {37},
  number = {1},
  pages = {1334--1342},
  year = {2023},
  doi = {10.1609/aaai.v37i1.25217}
}

@inproceedings{cho2024factorizephys,
  author = {Joshi, Jitesh and Agaian, Sos S. and Cho, Youngjun},
  title = {FactorizePhys: Matrix Factorization for Multidimensional Attention in Remote Physiological Sensing},
  booktitle = {Advances in Neural Information Processing Systems},
  volume = {37},
  pages = {96607--96639},
  year = {2024},
  doi = {10.52202/079017-3063}
}

@article{zou2025rhythmmamba,
  author = {Zou, Bochao and Guo, Zizheng and Hu, Xiaocheng and Ma, Huimin},
  title = {RhythmMamba: Fast, Lightweight, and Accurate Remote Physiological Measurement},
  journal = {Proceedings of the AAAI Conference on Artificial Intelligence},
  volume = {39},
  number = {10},
  pages = {11077--11085},
  year = {2025},
  doi = {10.1609/aaai.v39i10.33204}
}

@inproceedings{shao2025realworld,
  author = {Shao, Hang and Luo, Lei and Qian, Jianjun and Yan, Mengkai and Chen, Shuo and Yang, Jian},
  title = {Remote Photoplethysmography in Real-World and Extreme Lighting Scenarios},
  booktitle = {Proceedings of the IEEE/CVF Conference on Computer Vision and Pattern Recognition (CVPR)},
  pages = {10858--10867},
  year = {2025}
}

@inproceedings{zhao2026phasenet,
  author = {Zhao, Bo and Guo, Dan and Cao, Junzhe and Xu, Yong and Zou, Bochao and Tan, Tao and Sun, Yue and Yu, Zitong},
  title = {{PHASE-Net}: Physics-Grounded Harmonic Attention System for Efficient Remote Photoplethysmography Measurement},
  booktitle = {Proceedings of the IEEE/CVF Conference on Computer Vision and Pattern Recognition (CVPR)},
  pages = {21198--21207},
  year = {2026}
}

@inproceedings{dietterich2000ensemble,
  author = {Dietterich, Thomas G.},
  title = {Ensemble Methods in Machine Learning},
  booktitle = {Multiple Classifier Systems},
  pages = {1--15},
  year = {2000},
  publisher = {Springer},
  doi = {10.1007/3-540-45014-9_1}
}

@inproceedings{lakshminarayanan2017deep,
  author = {Lakshminarayanan, Balaji and Pritzel, Alexander and Blundell, Charles},
  title = {Simple and Scalable Predictive Uncertainty Estimation Using Deep Ensembles},
  booktitle = {Advances in Neural Information Processing Systems},
  volume = {30},
  year = {2017}
}

@misc{bai2025qwen3vl,
  author = {Bai, Shuai and others},
  title = {{Qwen3-VL} Technical Report},
  year = {2025},
  eprint = {2511.21631},
  archivePrefix = {arXiv},
  primaryClass = {cs.CV},
  url = {https://arxiv.org/abs/2511.21631},
  note = {arXiv:2511.21631}
}

@inproceedings{wei2022chain,
  author = {Wei, Jason and Wang, Xuezhi and Schuurmans, Dale and Bosma, Maarten and Ichter, Brian and Xia, Fei and Chi, Ed H. and Le, Quoc V. and Zhou, Denny},
  title = {Chain-of-Thought Prompting Elicits Reasoning in Large Language Models},
  booktitle = {Advances in Neural Information Processing Systems},
  volume = {35},
  pages = {24824--24837},
  year = {2022}
}

@inproceedings{wang2023selfconsistency,
  author = {Wang, Xuezhi and Wei, Jason and Schuurmans, Dale and Le, Quoc V. and Chi, Ed H. and Narang, Sharan and Chowdhery, Aakanksha and Zhou, Denny},
  title = {Self-Consistency Improves Chain of Thought Reasoning in Language Models},
  booktitle = {International Conference on Learning Representations},
  year = {2023}
}

@inproceedings{yao2023react,
  author = {Yao, Shunyu and Zhao, Jeffrey and Yu, Dian and Du, Nan and Shafran, Izhak and Narasimhan, Karthik and Cao, Yuan},
  title = {ReAct: Synergizing Reasoning and Acting in Language Models},
  booktitle = {International Conference on Learning Representations},
  year = {2023}
}

@inproceedings{yao2023tree,
  author = {Yao, Shunyu and Yu, Dian and Zhao, Jeffrey and Shafran, Izhak and Griffiths, Thomas L. and Cao, Yuan and Narasimhan, Karthik},
  title = {Tree of Thoughts: Deliberate Problem Solving with Large Language Models},
  booktitle = {Advances in Neural Information Processing Systems},
  volume = {36},
  pages = {11809--11822},
  year = {2023}
}

@inproceedings{du2024improving,
  author = {Du, Yilun and Li, Shuang and Torralba, Antonio and Tenenbaum, Joshua B. and Mordatch, Igor},
  title = {Improving Factuality and Reasoning in Language Models through Multiagent Debate},
  booktitle = {International Conference on Machine Learning},
  volume = {235},
  pages = {11733--11763},
  year = {2024}
}

@inproceedings{liu2023visual,
  author = {Liu, Haotian and Li, Chunyuan and Wu, Qingyang and Lee, Yong Jae},
  title = {Visual Instruction Tuning},
  booktitle = {Advances in Neural Information Processing Systems},
  volume = {36},
  pages = {34892--34916},
  year = {2023}
}

@inproceedings{zhao2020multiscale,
  author = {Zhao, Changchen and Han, Weiran and Chen, Zan and Li, Yongqiang and Feng, Yuanjing},
  title = {Remote Estimation of Heart Rate Based on Multi-Scale Facial {ROI}s},
  booktitle = {Proceedings of the IEEE/CVF Conference on Computer Vision and Pattern Recognition Workshops},
  pages = {278--279},
  year = {2020}
}

@article{song2025spectrogram,
  author = {Song, Rencheng and Du, Zhenzhou and Cheng, Juan and Li, Chang and Yang, Xuezhi},
  title = {Video-Based Heart Rate Estimation with Spectrogram Signal Quality Ranking and Fusion},
  journal = {Biomedical Signal Processing and Control},
  volume = {100},
  pages = {107094},
  year = {2025},
  doi = {10.1016/j.bspc.2024.107094}
}

@article{kiddle2023dynamic,
  author = {Kiddle, Adam and Barham, Helen and Wegerif, Simon and Petronzio, Connie},
  title = {Dynamic Region of Interest Selection in Remote Photoplethysmography: Proof-of-Concept Study},
  journal = {JMIR Formative Research},
  volume = {7},
  pages = {e44575},
  year = {2023},
  doi = {10.2196/44575}
}

@inproceedings{vanputten2023stacked,
  author = {{van Putten}, Lieke D. and Bamford, Kate E.},
  title = {Improving Systolic Blood Pressure Prediction From Remote Photoplethysmography Using a Stacked Ensemble Regressor},
  booktitle = {Proceedings of the IEEE/CVF Conference on Computer Vision and Pattern Recognition Workshops},
  pages = {5957--5964},
  year = {2023}
}

@article{perezgodoy2024desreg,
  author = {P{\'e}rez-Godoy, Mar{\'i}a D. and Molina, Marta and Mart{\'i}nez, Francisco and Elizondo, David and Charte, Francisco and Rivera, Antonio J.},
  title = {{DESReg}: Dynamic Ensemble Selection Library for Regression Tasks},
  journal = {Neurocomputing},
  volume = {580},
  pages = {127487},
  year = {2024},
  doi = {10.1016/j.neucom.2024.127487}
}

@article{joshi2024ibvp,
  author = {Joshi, Jitesh and Cho, Youngjun},
  title = {{iBVP} Dataset: {RGB-Thermal} {rPPG} Dataset with High Resolution Signal Quality Labels},
  journal = {Electronics},
  volume = {13},
  number = {7},
  pages = {1334},
  year = {2024},
  doi = {10.3390/electronics13071334}
}
}
\def\PhysAgentInlineAppendix{}
\appendix
\setcounter{equation}{0}
\setcounter{table}{0}
\setcounter{figure}{0}
\renewcommand{\theequation}{S\arabic{equation}}
\renewcommand{\thetable}{S\arabic{table}}
\renewcommand{\thefigure}{S\arabic{figure}}
\renewcommand{\theHequation}{supp.\arabic{equation}}
\renewcommand{\theHtable}{supp.\arabic{table}}
\renewcommand{\theHfigure}{supp.\arabic{figure}}
\section{Design Properties and Conditional Error Analysis}
\label{app:design_properties}

This appendix distinguishes properties enforced by the deterministic
decision--verification pipeline from empirical cluster-selection accuracy.
The analysis does not assume that the Judge Agent is optimal, nor does it
claim that access to additional agent evidence must monotonically improve
heart-rate accuracy. Instead, it characterizes what can be guaranteed once a
decision satisfies the verifier's executable, single-cluster contract with
finite nonnegative normalized weights. The
empirical value of the additional evidence is evaluated by the ablations in
Table~\ref{tab:ablation_results} and the cases in
Appendix~\ref{app:case_studies}.

\paragraph{Setup.}
Let \(\mathcal{C}_{\rm v}\) be the nonempty set of valid candidates and let
\(\mathcal{G}=\{G_j\}_{j=1}^{J}\) be the hypothesis clusters constructed by
the deterministic clustering routine before the Judge step. For each realized
cluster \(G_j\), define
\begin{equation}
\ell_j=\min_{c_i\in G_j}h_i,\qquad
u_j=\max_{c_i\in G_j}h_i,\qquad
D_j=u_j-\ell_j,
\end{equation}
and
\begin{equation}
\mu_j=\frac{1}{|G_j|}\sum_{c_i\in G_j}h_i,\qquad
r_j=\max_{c_i\in G_j}|h_i-\mu_j|.
\end{equation}
Here, \(D_j\) and \(r_j\) are measured after clustering, and
\(D_j\leq 2r_j\). The clustering tolerance \(\epsilon\) controls candidate
admission during cluster construction; it is not assumed to be either the
final cluster radius or an error bound relative to the unknown ground truth.

The Judge Agent receives both agent-organized evidence and deterministic
physiological evidence and produces
\begin{equation}
d_{\rm MLLM}
=A_{\rm judge}(\mathcal{C},z,\tau,b,\mathcal{G},E,\mathcal{H}).
\end{equation}
The decision may specify a strategy, selected candidates, exclusions, and
weights, but it is not itself a heart-rate estimate. The verifier and fuser
produce
\begin{equation}
d^*=V(d_{\rm MLLM},\mathcal{G},E,\mathcal{H}),\qquad
\hat{y}=F(d^*,\mathcal{C}).
\end{equation}

\paragraph{Property 1 (decision--output separation).}
For every executable PhysAgent output, the scalar \(\hat{y}\) is computed by
the deterministic function \(F\) after decision verification. Consequently,
the MLLM agents can influence which hypothesis and fusion parameters are
proposed, but no free-form numerical value emitted by an agent is used
directly as the final heart rate. This is an implementation property and does
not imply that the selected hypothesis is correct. This property characterizes
how a given decision is executed and does not assume that the decision itself
remains identical across repeated agent runs.

\paragraph{Verifier-covered domain.}
We call an execution verifier-covered when PCHV is enabled, valid nonempty
cluster evidence assigns every executable candidate to a cluster, and the
fuser operates on the nonempty set returned by the verifier without invoking
a no-evidence or empty-set last resort. On this domain, an admissible decision
is accepted only after the single-cluster, support, and detected-harmonic
checks. An inadmissible decision is repaired to one admissible cluster, while
an unavailable or unparsable decision invokes cluster-based fallback.
Therefore, for the effective selected set \(S^*\), the accepted, repaired, and
fallback branches all satisfy
\begin{equation}
S^*\subseteq G_{j^*}.
\end{equation}
This statement is relative to the computed cluster evidence; it does not
assert that \(G_{j^*}\) is the ground-truth cluster. Degenerate inputs without
valid cluster evidence or without a nonempty executable set are outside this
single-cluster statement and cannot be used to derive an accuracy guarantee.

\paragraph{Property 2 (within-hypothesis containment).}
For a verifier-covered decision, single selection, median, trimmed mean, and
a weighted mean with \(w_i\geq 0\) and
\(\sum_{c_i\in S^*}w_i=1\) are all range preserving. Hence,
\begin{equation}
\ell_{j^*}\leq \hat{y}\leq u_{j^*},\qquad
|\hat{y}-\mu_{j^*}|\leq r_{j^*}.
\end{equation}
For single selection the output is a member of \(S^*\); median and trimmed
mean remain between the smallest and largest selected values; and the
normalized weighted mean lies in their convex hull. Since
\(S^*\subseteq G_{j^*}\), all four outputs lie within the realized cluster
range. Thus, within the verifier-covered domain, PCHV prevents an intermediate
estimate formed by averaging candidates from different hypothesis clusters,
although it may still select an incorrect cluster.

\paragraph{Property 3 (bounded influence of fusion weights).}
Consider two normalized nonnegative weight vectors \(w\) and \(w'\) supported
on the same cluster \(G_j\). Their weighted estimates satisfy
\begin{equation}
\left|\hat{y}(w)-\hat{y}(w')\right|
\leq \frac{D_j}{2}\|w-w'\|_1
\leq D_j.
\end{equation}
To see this, let \(m_j=(\ell_j+u_j)/2\). Because
\(\sum_i(w_i-w'_i)=0\),
\begin{equation}
\begin{aligned}
\left|\sum_i(w_i-w'_i)h_i\right|
&=
\left|\sum_i(w_i-w'_i)(h_i-m_j)\right|\\
&\leq \frac{D_j}{2}\|w-w'\|_1.
\end{aligned}
\end{equation}
Thus, cluster verification bounds, but does not eliminate, the numerical
influence of fusion weights: changing weights cannot move the estimate
outside the realized hypothesis range. The bound does not cover selecting a
different cluster.

\paragraph{Conditional error decomposition.}
For any ground-truth heart rate \(y\), Property 2 yields
\begin{equation}
|\hat{y}-y|
\leq |\mu_{j^*}-y|+r_{j^*}.
\end{equation}
If \(G_{j^*}\) contains a candidate \(h_a\) satisfying
\(|h_a-y|\leq\delta\), then both \(\hat{y}\) and \(h_a\) lie in the same
cluster range, giving
\begin{equation}
|\hat{y}-y|
\leq \delta+D_{j^*}
\leq \delta+2r_{j^*}.
\end{equation}

\begin{figure*}[!t]
    \centering
    \includegraphics[width=\textwidth]{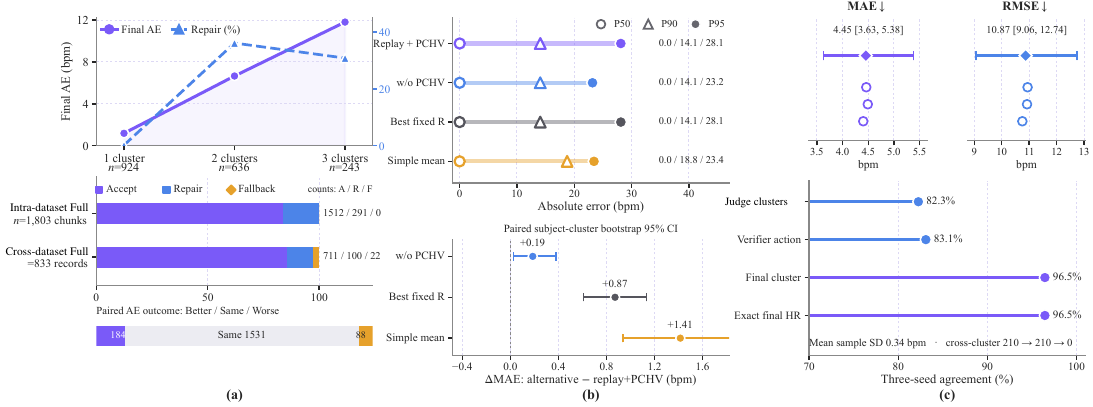}
    \caption{Aggregate PhysAgent reliability evidence from complementary
    frozen audit protocols: (a) candidate conflict and PCHV actions,
    (b) paired recoverability under identical frozen decisions, and
    (c) final-output consistency across three seeds on a fixed,
    dataset-balanced random sample of 479 cases drawn once before repeated
    inference.}
    \label{fig:reliability_evidence}
\end{figure*}

Accordingly,
\begin{equation}
\mathbb{E}[|\hat{y}-y|]
\leq
\mathbb{E}[|\mu_{j^*}-y|]
+
\mathbb{E}[r_{j^*}].
\end{equation}
The first term reflects which cluster is selected and must be established
empirically; the second records within-cluster dispersion. This decomposition
does not infer improved selection accuracy merely from the Judge Agent having
access to more evidence.

\subsection{Empirical Reliability, Recoverability, and Repeatability}
\label{app:empirical_reliability}

Figure~\ref{fig:reliability_evidence} evaluates PhysAgent from three
complementary perspectives: candidate conflict, paired recovery, and
multi-seed output stability. To isolate the deterministic contribution of
PCHV, the paired analysis replays the same frozen decision with and without
PCHV, without regenerating the upstream MLLM decision. Full-run action counts
are reported separately for the \reviewchange{intra-dataset} and \reviewchange{cross-dataset} protocols so that
their different evaluation units and candidate pools remain explicit. These
analyses provide mechanism-level reliability evidence and do not replace the
headline end-to-end evaluation in the main tables.

\paragraph{Targeted PCHV intervention.}
In the controlled cohort of 1,803 samples, 924, 636, and 243 samples contain
one, two, and three candidate HR clusters, respectively; their mean final
absolute errors are 1.19, 6.66, and 11.81 bpm. The number of candidate
clusters is positively associated with final error, indicating that
multi-hypothesis conflict characterizes the more difficult samples. Of the
307 repairs in this replay, 305 occur on multi-cluster samples, showing that
PCHV intervention is highly concentrated on genuine candidate conflicts
rather than on already consistent predictions.

The complete end-to-end logs use two separate reporting units. The \reviewchange{intra-dataset}
Full run contains 1,803 chunks, with Accept/Repair/Fallback counts of
1,512/291/0. The \reviewchange{cross-dataset} Full run in
Table~\ref{tab:cross_dataset_leave_one_out} contains 833 unique target records,
with counts of 711/100/22.

The paired outcome bar is not a decomposition of these full-run action bars;
it is obtained from the same-decision controlled replay. PCHV leaves the
absolute error unchanged in 1,531 cases, or 84.9\% of the cohort,
demonstrating conservative behavior. Among the 272 cases whose absolute error
changes, 184 improve and 88 worsen: beneficial changes are approximately 2.1
times as frequent and account for 67.6\% of changed cases. The overall MAE
decreases from 4.7354 to 4.5495 bpm. Thus, PCHV passes most structurally
admissible decisions without modification and, when it changes the output, is
more likely to correct than to harm it.

\paragraph{Average recoverability.}
For the controlled replay, we compare PCHV with the same decision executed
without PCHV, the best fixed candidate RhythmFormer, and simple averaging. We
define
\[
\Delta\mathrm{MAE}
=
\mathrm{MAE}_{\mathrm{alternative}}
-
\mathrm{MAE}_{\mathrm{replay+PCHV}},
\]
so a positive value favors PCHV. The paired
\(\Delta\mathrm{MAE}\) values are \(+0.19\), \(+0.87\), and \(+1.41\) bpm,
with 95\% confidence intervals of \([0.03,0.38]\), \([0.61,1.13]\), and
\([0.94,1.92]\), respectively. Hence, after holding the decision fixed, PCHV
provides a consistent average-MAE advantage over removing verification,
always selecting the same candidate, or directly averaging the candidates.
Some difficult high-error cases still exert substantial influence on RMSE
and upper-tail quantiles, so this result supports improved average
recoverability rather than uniform dominance at every tail quantile. These
remaining cases provide a concrete target for further calibration of repair
conditions and cluster selection.

\paragraph{MLLM repeatability and final-output stability.}
We draw one dataset-balanced random sample of 479 cases before any repeated
inference or result inspection, freeze the sample identities without
outcome-based selection, and then repeat the full inference procedure on the
identical cohort with three independent random seeds, yielding 1,437
decisions. The mean MAE is 4.45 bpm with a 95\%
confidence interval of \([3.63,5.38]\), and the mean RMSE is 10.87 bpm with
\([9.06,12.74]\). These intervals are obtained by a paired hierarchical
bootstrap over samples and repeated runs, rather than from the three seed
means alone.

The Judge-selected cluster and verifier action are identical across all three runs
for 82.3\% and 83.1\% of the fixed cases, respectively. After verification
and deterministic fusion, both the final-cluster agreement and exact
final-HR agreement reach 96.5\%, while the mean sample-level standard
deviation of the final HR is only 0.34 bpm. Moreover, all 210 cross-cluster
fusion decisions in the audit are repaired, leaving no residual
cross-cluster execution. Thus, upstream MLLM decisions may retain some
stochasticity, but that variation rarely propagates to the final measurement.
Reliability here does not require identical intermediate reasoning on every
run; PCHV and deterministic fusion constrain its effect on the final HR.

Overall, the audit shows that PCHV concentrates intervention on conflicting
samples, passes most structurally admissible decisions without modification,
and is more often beneficial than harmful when it changes the output. Together
with the high final-output agreement, these results support the reliability
and auditability of PhysAgent under candidate conflict and upstream
variability.

\FloatBarrier

\section{Datasets and Evaluation Protocols}
\label{app:datasets_protocols}

This appendix specifies the data and evaluation protocols used in
Section~4. PhysAgent is an inference-time candidate-verification method:
it does not train a new rPPG backbone in the reported experiments, but
instead consumes saved candidate heart-rate estimates and waveforms from
base rPPG models. The role of this appendix is therefore to make clear
which clips are evaluated, how the base models are obtained, which
models provide candidate hypotheses, and how the aggregate metrics in the
main tables are computed. We additionally report the token usage and
inference cost of the agent backbone.

\subsection{\reviewchange{Datasets}}

We evaluate on four public rPPG benchmarks: UBFC-rPPG, PURE, BUAA-MIHR, and
MMPD. All
heart-rate values are reported in beats per minute (bpm). The experiments
use processed clips and synchronized pulse/BVP labels from the rPPG
evaluation pipeline. Candidate waveforms use 128-frame, 30-fps chunks; cross-dataset
HRs and targets are extracted after concatenating chunks by video. In the main \reviewchange{intra-dataset} evaluation, BUAA-MIHR uses the
0.0--0.72 partition for training and the 0.72--1.0 partition for testing,
while MMPD uses the 0.0--0.7 partition for training and the 0.8--1.0
partition for testing. For PURE and UBFC-rPPG, we use the publicly released
official model checkpoints and evaluate them on the 0.9--1.0 and
0.72--1.0 test partitions, respectively; their official training
partitions are 0.0--0.6 and 0.0--0.72, respectively. The held-out test
sets contain 177 chunks for UBFC-rPPG, 95 chunks for PURE, 336 chunks for
BUAA-MIHR, and 1195 chunks for MMPD. PhysAgent and the compared base
models are evaluated on the same test chunks within each protocol.

\subsection{Base-Model Sources and Candidate Boundaries}
\label{app:base_model_sources}

For each base rPPG model, we use its public checkpoint when it is available
and compatible with our evaluation grid; otherwise, we retrain it following the
original protocol, independently on one NVIDIA H100 GPU.

Because the \reviewchange{intra-dataset} and cross-dataset protocols impose different checkpoint
availability and training constraints, we use protocol-specific candidate
pools. For the \reviewchange{intra-dataset} evaluation, the PhysAgent candidate set
\(\mathcal{C}\) contains aligned outputs from RhythmFormer, PhysMamba, and
FactorizePhys. For the cross-dataset evaluation, it contains outputs from
source-trained PhysFormer, EfficientPhys, and RhythmFormer. These three models
can be trained consistently using only source-domain data under every
leave-one-out and limited-source split. Within each protocol, PhysAgent and all
conventional fusion baselines consume exactly the same precomputed candidate
outputs. Each pool is fixed before evaluation on the corresponding test split,
and target-domain test labels are never used for model selection, candidate
construction, or fusion-rule tuning. The change in backbone identities therefore
reflects protocol-specific source-training requirements rather than
target-performance-driven selection. PhysNet, PhysMamba, the classical
methods, and all other table rows are comparison baselines rather than
additional PhysAgent candidates.

Figure~\ref{fig:sensitivity_analysis}(a) uses the following abbreviations:
P denotes PhysNet, R denotes RhythmFormer, M denotes PhysMamba, F denotes
FactorizePhys, and I denotes iBVPNet \citep{joshi2024ibvp}. The three-model \(R+M+F\) reference uses
the latest unrounded PhysAgent averages reported in
Table~\ref{tab:intra_dataset_results}.

\subsection{Conventional Multi-Model Fusion Baselines}
\label{app:conventional_fusion}

Within each protocol, the five baselines use exactly the same aligned candidate
outputs as PhysAgent. For the intra-dataset evaluation in
Table~\ref{tab:intra_dataset_results}, the shared pool consists of RhythmFormer,
PhysMamba, and FactorizePhys. For the cross-dataset evaluations in
Tables~\ref{tab:cross_dataset_leave_one_out},~\ref{tab:cross_dataset_limited_mmpd},
and~\ref{tab:cross_dataset_limited_buaa}, the shared pool consists of the
source-trained outputs of PhysFormer, EfficientPhys, and RhythmFormer. We adapt
the cited aggregation principles to each common candidate pool rather than
claim exact reproduction of the original pipelines. The tables abbreviate
these adaptations as follows:
\textbf{Uniform Mean} averages the three HR estimates, inspired by equal-weight
multi-scale ROI fusion \citep{zhao2020multiscale}; \textbf{SQI Weighting} uses
a composite signal-quality score, inspired by signal-quality ranking and fusion
\citep{song2025spectrogram}; \textbf{Max-SNR} selects the highest-SNR waveform,
inspired by SNR-guided ROI selection and aggregation \citep{kiddle2023dynamic};
\textbf{Ridge Stacking} learns a global linear combination, inspired by stacked
regression for blood-pressure prediction \citep{vanputten2023stacked}; and
\textbf{Local Competence} adapts dynamic ensemble selection for regression
\citep{perezgodoy2024desreg} using local validation error in a
\(k\)-nearest-neighbor competence region.

The first three rules have no learned parameters. Ridge regularization and
\(k\) are selected on validation data, after which train+validation is used for
fitting or competence estimation. In the cross-dataset setting, these steps use
only source-domain train/validation outputs. Target-domain labels are used only
to compute final metrics, never to fit or tune a fusion rule.

\subsection{Intra-Dataset Evaluation}

In the intra-dataset protocol, each dataset is evaluated on its held-out
test split. PhysAgent uses only the candidate outputs produced by the
base models before metric computation. No ground-truth heart rate is
used during candidate construction, candidate selection, or final
prediction.

For Table~\ref{tab:intra_dataset_results}, MAE, RMSE, and Pearson
correlation \(R\) are first computed separately for each dataset. The average
MAE and RMSE entries are sample-weighted averages:
\begin{equation}
\bar{m}_{w}=\frac{\sum_d n_d m_d}{\sum_d n_d},
\qquad m\in\{\mathrm{MAE},\mathrm{RMSE}\},
\end{equation}
where \(d\) indexes datasets, \(n_d\) is the number of test chunks, and
\(m_d\) is the corresponding dataset-level metric. The average \(R\) is the
arithmetic mean of the four dataset-level correlations:
\begin{equation}
\overline{R}=\frac{1}{D}\sum_{d=1}^{D}R_d,\qquad D=4.
\end{equation}
The average \(R\) is a descriptive dataset-level summary rather than Pearson
correlation recomputed after pooling samples from different datasets. It
is omitted when any dataset-level \(R\) is unavailable. For corrected fusion RMSEs and UBFC-updated rows, aggregates combine
recomputed values with the remaining reported dataset metrics; other
aggregates use unrounded values.

\subsection{Cross-Dataset Evaluation}

For cross-dataset evaluation, we use the notation \(U=\)UBFC-rPPG,
\(P=\)PURE, \(B=\)BUAA-MIHR, and \(M=\)MMPD. In the leave-one-out protocol,
one dataset is held out as the target domain and all remaining datasets
serve as source domains:
\[
\begin{aligned}
P+B+M&\rightarrow U, &
U+B+M&\rightarrow P,\\
U+P+M&\rightarrow B, &
U+P+B&\rightarrow M .
\end{aligned}
\]
In the limited-source protocol, only two source domains are available:
\[
\begin{aligned}
P+B&\rightarrow M, &
P+U&\rightarrow M, &
B+U&\rightarrow M,
\end{aligned}
\]
and
\[
\begin{aligned}
P+M&\rightarrow B, &
M+U&\rightarrow B, &
P+U&\rightarrow B .
\end{aligned}
\]

For these protocols, target-domain labels are used only after prediction
for computing final metrics. They are not used for base-model
training, validation, early stopping, candidate construction, candidate
selection, or fusion-rule tuning.

The average columns in the cross-dataset tables report arithmetic means
of MAE, RMSE, and \(R\) over the corresponding transfer protocols: four
target domains for the leave-one-out table and three source-pair protocols
for each limited-source table. The average \(R\) therefore summarizes
protocol-level correlations rather than a correlation recomputed from
pooled target-domain samples. Lower MAE/RMSE and higher \(R\) indicate
better heart-rate estimation.

\subsection{Component Ablation Replacements and Rules-Only Control}
\label{app:ablation_replacements}

This subsection specifies how the archived component configurations replace
removed modules rather than leaving their downstream inputs undefined.
The component variants share the saved base-model outputs within each
archived dataset or transfer protocol. Removing a module changes its
replacement input and the downstream decisions; it does not reuse a final
Full prediction as the ablation output.

\paragraph{Without Scene Agent.}
The scene-model call is replaced by fixed defaults: low motion, even
illumination, unknown skin appearance, and no occlusion. All method priors
are set to \(0.5\), visual-evidence strength is low, and scene memory is not
used. Quality, Debate, Judge, and decision verification continue with these
inputs. The scene-conditioned deterministic machinery remains active with
the defaults. Thus, this configuration removes clip-specific visual scene
assessment rather than deleting the scene interface or all scene-conditioned
arithmetic.

\paragraph{Without Quality Agent.}
The language-model trust assessment is replaced by the candidate's raw
waveform-quality score \(u_i\), clipped to \([0,1]\); failed candidates
receive zero trust. Deterministic quality flags are retained, and subsequent
agents and the verifier consume these replacement rankings. Scene perception,
debate, and decision generation otherwise remain active.

\paragraph{Without Debate Agent.}
The Proponent and Skeptic calls are skipped and their output is replaced by
an empty opinion list. The Judge still receives candidates, scene tags,
quality rankings, and precomputed cluster and harmonic evidence before
decision verification and numerical fusion.

\paragraph{Deterministic replacement of the MLLM Judge.}
The archived \texttt{deterministic} mode selects a hypothesis using fixed
signal-quality, cluster-support, scene, and harmonic rules. It then returns
single selection or signal-quality-weighted fusion within the selected
cluster. Scene perception remains active. This mode also replaces Quality's
language-model trust with signal-quality rankings and skips Debate, because
the rule selector does not use their generated outputs. It therefore tests a
joint replacement of the language-based decision path, not an intervention
on the Judge alone. Its own rule-based harmonic handling replaces the
fusion-decision verification path.

\paragraph{Without PCHV.}
The Judge retains candidate, scene, trust, debate, cluster, and harmonic
inputs, but an executable decision is passed directly to the numerical fuser
without the normal decision verification and repair. Parsing and
fallback for unavailable decisions remain available. Cluster and harmonic evidence
are still visible to the Judge, so this intervention removes downstream
verification rather than the upstream physiological evidence.

\paragraph{Without all MLLM agents (rules only).}
This control uses no generated scene tags, method priors, trust assessments,
debate opinions, or Judge decision. On the saved candidates, trust is replaced
by \(\tau_i=\operatorname{clip}_{[0,1]}(u_i)\), scene support is neutral,
and adaptive scene diagnostics are omitted. The fixed cluster-support rule
selects a group and forms a deterministic decision, which is checked by the
archived verifier with its harmonic rules. A singleton returns its candidate
HR; a multi-member group uses normalized \(\max(u_i,0.05)\) weights.
All final arithmetic uses the saved numerical candidate HRs. No model calls,
new fitting, or outcome-based rule selection are involved in this replay.

\paragraph{Scope and aggregation of the rules-only result.}
The separate row in Table~\ref{tab:ablation_results} is recomputed from the
historical component-ablation candidate archive. Both archived \reviewchange{intra-dataset} and
\reviewchange{cross-dataset} controls use RhythmFormer, PhysMamba, and FactorizePhys.
\reviewchange{Intra-dataset} MAE and RMSE are computed after pooling all \(1{,}803\) test
chunks (177 UBFC-rPPG, 95 PURE, 336 BUAA-MIHR, and 1,195 MMPD). This pooled
RMSE differs from the sample-weighted mean of dataset RMSEs used in
Table~\ref{tab:intra_dataset_results}. \reviewchange{Cross-dataset} evaluation uses
\(833\) video-level records: 42, 59, 72, and 660 for
Others\(\rightarrow\)U/P/B/M, respectively; the four target-domain metrics
are averaged equally. These computations give 5.30/11.15 and 7.32/12.53 bpm.

The archived \reviewchange{cross-dataset} control pool differs from the
PhysFormer/EfficientPhys/RhythmFormer pool described for the current main
cross-dataset tables. Moreover, the archived Table~5 protocol imports the
reported Full averages rather than rerunning Full alongside its component
variants; a matching raw Full run for the current references has not been
established here. \reviewchange{Parenthesized percentages for the rules-only row are arithmetic changes from the unrounded recomputed errors relative to the reported Full averages. The row retains the archived candidate pools and aggregation defined above.} The replay implementation reproduces all \(1{,}803\)
previously audited signal-only predictions to numerical precision. Using
only \(2\times\) instead of \(\{2,3\}\) harmonic checks leaves the newly
reconstructed control outputs unchanged. These are source and replay checks,
not independent performance validation.

\subsection{Token Usage and Inference Cost}
\label{app:token_cost}

We measure the token usage and computational cost of the Qwen3-VL-4B-based
agent stage in PhysAgent. The model is deployed locally on one NVIDIA RTX
4090 GPU and therefore incurs no external token-based API fees. Each sample
uses an average of 18.45K tokens and requires 12.589 seconds of serial
wall-clock time. With eight-way concurrency and an assumed GPU rental price
of RMB~0.8 per hour, this corresponds to an amortized processing time of
approximately 1.574 seconds and an estimated compute cost of approximately
RMB~0.00035 per sample. These values are theoretical estimates under ideal
concurrent utilization; actual end-to-end costs may be slightly higher due
to task scheduling, incomplete hardware utilization, batching overhead, and
retries for structured output.

\section{Case-Study Details}
\label{app:case_studies}

\subsection{Selection and Evaluation Setting}

\begin{figure*}[t]
    \centering
    \includegraphics[width=\textwidth]{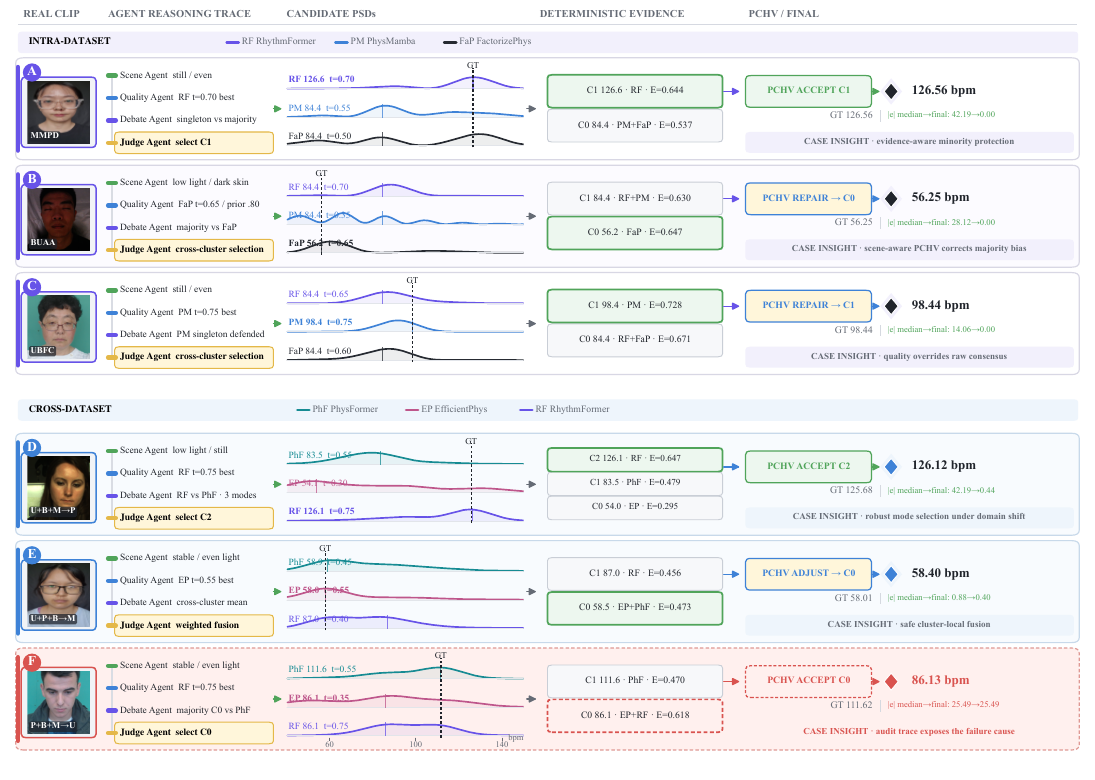}
    \caption{Representative protocol-matched PhysAgent traces: Rows A--C are
    \reviewchange{intra-dataset}, Rows D--F are \reviewchange{cross-dataset}, and Row F is a failure case.}
    \label{fig:protocol_case_gallery}
\end{figure*}

Figure~\ref{fig:protocol_case_gallery} presents six representative cases from
the frozen inference records. Cases A--C use the \reviewchange{intra-dataset} candidate pool of
RhythmFormer (RF), PhysMamba (PM), and FactorizePhys (FaP), whereas Cases D--F
use the \reviewchange{cross-dataset} candidate pool of PhysFormer (PhF), EfficientPhys (EP),
and RhythmFormer (RF). The cases were selected after the experiments to
illustrate complementary mechanisms, including candidate-conflict
organization, evidence-aware minority protection, cross-cluster decision
repair, safe cluster-local fusion, and failure auditing. Ground-truth heart
rate is used only for post-hoc analysis and visualization; it is never
available to any agent, the Judge, or PCHV during inference.

\subsection{Mechanism-Level Interpretation}

\paragraph{Case A: evidence-aware minority protection.}
PM and FaP form the majority cluster C0 at 84.38 bpm, while RF alone forms C1
at 126.56 bpm. The Quality Agent assigns RF the highest trust, the Judge
selects C1, and PCHV accepts the decision. This case shows that PhysAgent can
protect a minority hypothesis when it is better supported by the evidence,
rather than simply following a model majority.

\paragraph{Case B: scene-conditioned correction of majority bias.}
RF and PM form the majority cluster C1 at 84.38 bpm, while FaP alone forms C0
at 56.25 bpm. The Scene Agent detects low illumination and darker skin
appearance, and the scene-conditioned and quality evidence keeps the FaP
hypothesis supported. The Judge produces a cross-cluster decision, which PCHV
repairs to the better-supported C0. This case illustrates scene-conditioned
reliability assessment and correction of an erroneous majority bias.

\paragraph{Case C: candidate-specific quality overrides superficial consensus.}
RF and FaP form the majority cluster C0 at 84.38 bpm, while PM alone forms C1
at 98.44 bpm. PM receives the highest trust, and the Debate Agent preserves
the singleton hypothesis. After the Judge produces a cross-cluster decision,
PCHV repairs it to C1. Thus, candidate-specific signal quality can allow a
reliable singleton to prevail over apparent model consensus.

\paragraph{Case D: robust mode selection under domain shift.}
In this \reviewchange{cross-dataset} case, EP, PhF, and RF form three distinct HR modes. The
Scene Agent identifies low illumination, the Quality Agent assigns RF the
highest trust, and the Judge selects RF's cluster C2; PCHV accepts this
decision. The case shows that PhysAgent can select an evidence-supported mode
under substantial domain shift instead of averaging mutually incompatible
predictions.

\paragraph{Case E: safe cluster-local fusion.}
EP and PhF form a consistent C0 near 58 bpm, while RF alone forms C1 at
87.01 bpm. The Debate Agent flags the risk of a cross-cluster average, and the
Judge issues a weighted decision. PCHV constrains the fusion to C0 and produces
58.40 bpm. This case demonstrates safe within-cluster fusion that avoids an
unsupported intermediate HR.

\paragraph{Case F: an auditable failure.}
EP and RF form a high-trust majority cluster C0 at 86.13 bpm, whereas the
correct PhF prediction at 111.62 bpm is the singleton C1. The Judge selects C0,
and PCHV accepts it under the available evidence, yielding an incorrect final
result. This case is not presented as evidence of an accuracy gain; instead,
the complete trace exposes how the supported majority suppressed the correct
singleton and makes the failure auditable.

Together, these cases clarify the division of labor in PhysAgent: the agents
and Judge organize evidence and propose candidate hypotheses, while PCHV
enforces executable single-cluster and physiological consistency. Case F also
shows the boundary of that guarantee: PCHV ensures structural executability
and auditability, but it cannot guarantee that the selected cluster is
correct.

\section{Signal-Quality Feature Definitions}
\label{app:signal_quality}

PhysAgent computes these features deterministically and uniformly across
candidate waveforms, without an MLLM, scene tags, or ground-truth heart rate.

\paragraph{Spectrum-based features.}
For a length-\(N\) waveform sampled at frame rate \(f\), we mean-center it as
\(\widetilde{s}_i[n]=s_i[n]-N^{-1}\sum_{m=1}^{N}s_i[m]\) and estimate its
power spectral density \(p_i(k)\) at frequency \(\nu_k\) using Welch's method,
with segment length
\(N_{\rm seg}=\min(N,\max(\lfloor4f\rfloor,64))\). We use the plausible
heart-rate band
\(\mathcal{B}=\{k:0.7\leq\nu_k\leq3.5~\mathrm{Hz}\}\), corresponding to
\(42\)--\(210\) bpm. Let
\[
\begin{aligned}
P_i^{\rm band}&=\sum_{k\in\mathcal{B}}p_i(k),&
P_i^{\rm peak}&=\max_{k\in\mathcal{B}}p_i(k),\\
P_i^{\rm rest}&=P_i^{\rm band}-P_i^{\rm peak},&
\overline{P}_i^{\rm band}&=P_i^{\rm band}/|\mathcal{B}|.
\end{aligned}
\]
With \(\varepsilon=10^{-12}\) and
\(r_i=P_i^{\rm peak}/P_i^{\rm rest}\), the SNR is
\begin{equation}
\mathrm{SNR}^{\rm dB}_i=
\begin{cases}
-30, & P_i^{\rm band}<\varepsilon,\\
30, & P_i^{\rm rest}<\varepsilon,\\
10\log_{10}\!\left[\max(r_i,10^{-6})\right],
& \text{otherwise}.
\end{cases}
\end{equation}
Its normalized value and the normalized peak prominence are
\begin{equation}
\begin{aligned}
\widehat{\mathrm{SNR}}_i
&=\left[1+\exp(-\mathrm{SNR}^{\rm dB}_i/5)\right]^{-1},\\
P_i&=\min\!\left(
\frac{P_i^{\rm peak}}
{P_i^{\rm peak}+\overline{P}_i^{\rm band}+\varepsilon},1\right).
\end{aligned}
\end{equation}
For \(\widetilde{p}_i(k)=p_i(k)/P_i^{\rm band}\), let \(K_i\) be the number
of non-negligible in-band bins. The normalized spectral entropy and
out-of-band energy ratio are
\begin{equation}
\begin{aligned}
H_i&=-\frac{\sum_{k\in\mathcal{B}}
\widetilde{p}_i(k)\log\widetilde{p}_i(k)}{\log K_i},\\
M_i&=\operatorname{clip}_{[0,1]}\!\left(
1-\frac{P_i^{\rm band}}{\sum_kp_i(k)}\right).
\end{aligned}
\end{equation}
We set \(H_i=0\) if fewer than two non-negligible bins remain. \(M_i\) is an
out-of-band energy proxy, distinct from the Scene Agent's frame-level motion.

\paragraph{Temporal periodicity.}
Let
\(u_i[n]=\widetilde{s}_i[n]/(\operatorname{std}(\widetilde{s}_i)+\varepsilon)\).
We compute
\begin{equation}
\begin{aligned}
\rho_i(\ell)
&=\frac{\sum_{n=1}^{N-\ell}u_i[n]u_i[n+\ell]}
{\sum_{n=1}^{N}u_i[n]^2},\\
\ell_{\min}&=\max(\lfloor f/3.6\rfloor,1),\\
\ell_{\max}&=\min(\lfloor f/0.6\rfloor,\lfloor N/2\rfloor).
\end{aligned}
\end{equation}
If \(\mathcal{L}_i\) is the set of autocorrelation local-peak lags in this
range, the periodicity is
\begin{equation}
R_i=\operatorname{clip}_{[0,1]}
\left(\max_{\ell\in\mathcal{L}_i}\rho_i(\ell)\right).
\end{equation}
When no local peak is detected, we use the maximum autocorrelation in the
valid lag range. These definitions yield the all-good-is-high vector
\(\phi_i=[\widehat{\mathrm{SNR}}_i,P_i,1-H_i,1-M_i,R_i]\) used by the
scene-conditioned reliability module.

\section{Scene Mapping and Fusion Decision Execution Rules}
\label{app:implementation_details}

\paragraph{Weight selection.}
The numerical weighting configuration reported below was selected through
multiple rounds of experimental comparison. We retained the best-performing
configuration among those examined and fixed it for inference. The
signal-reliability considerations below explain the roles of the terms and
the directions of scene modulation; the numerical magnitudes were selected
empirically.

\paragraph{Fixed scene mapping.}
The vector order is
\(s(z)=(s_{\rm motion},s_{\rm illumination},s_{\rm occlusion},s_{\rm appearance})^\top\).
Table~\ref{tab:scene_risk_mapping} specifies the categorical encoding. Tags
are lowercased; missing or unrecognized entries use the defaults. The
appearance entry is a routing coefficient for a perceived visual tag, not
a calibrated physiological measurement.

\begin{table}[t]
\centering
\small
\caption{Scene-to-risk mapping. Defaults apply to missing or unrecognized tags.}
\label{tab:scene_risk_mapping}
\setlength{\tabcolsep}{3pt}
\begin{tabular}{@{}p{0.19\linewidth}p{0.61\linewidth}r@{}}
\hline
Risk & Tag: value & Default\\
\hline
Motion & still: 0; low: 0.25; moderate: 0.65; high: 1 & 0.25\\
Illumination & even: 0; low: 0.90; harsh: 0.85; flicker: 1 & 0\\
Occlusion & none: 0; partial: 0.60; heavy: 1 & 0\\
Appearance & fair: 0; medium: 0.10; dark: 0.80; unknown: 0.50 & 0.50\\
\hline
\end{tabular}
\end{table}

The rows of \(B\) follow
\((\widehat{\mathrm{SNR}},P,1-H,1-M,R)\); its columns follow \(s(z)\):
\begin{equation}
B=\begin{bmatrix}
-0.20 & 0.36 & -0.08 & 0.18\\
-0.10 & 0.28 &  0.06 & 0.16\\
-0.10 & 0.34 &  0.04 & 0.18\\
 0.38 &-0.20 &  0.24 &-0.08\\
 0.34 &-0.10 &  0.20 & 0.02
\end{bmatrix}.
\end{equation}
The base vector, in the same feature order, is
\begin{equation}
\alpha^0=(0.24,0.20,0.18,0.18,0.20)^\top.
\end{equation}
Both \(B\) and \(\alpha^0\) are held fixed during inference.
Their roles differ: \(\alpha^0\) includes all five
complementary waveform diagnostics, while \(B\) changes their relative
emphasis under scene risks. The matrix increases out-of-band rejection and
periodicity weights under motion or occlusion, and spectral-concentration
weights under adverse illumination. These directions reflect qualitative
signal-reliability considerations; the coefficient magnitudes belong to
the experimentally selected weighting configuration described above.
Softmax ensures positive weights summing to one, and yields \(\alpha^0\)
when \(s(z)=0\).

Visual-evidence strength maps low/medium/high to \(0.35/0.65/0.90\), with
default \(0.65\) and a cap of \(0.50\) for heuristic scene fallback.
Candidate disagreement is the valid HR span divided by \(40\) bpm, clipped
to \([0,1]\), and is zero for fewer than two candidates. Priors \(v_i\)
in the adaptive diagnostic use parsed values clipped to \([0,1]\), with
method-profile defaults archived in \texttt{reproducibility/agent\_prompts.txt}.
In contrast, the cluster support uses parsed priors \(p_i\), defaulting to
\(0.5\) for a missing method. This distinction follows the implementation.

\paragraph{Raw quality and contextual support.}
The scene-independent aggregate is \(u_i=\gamma^\top\phi_i\), with
\begin{equation}
\gamma=(0.30,0.20,0.15,0.15,0.20)^\top.
\end{equation}
Degenerate waveforms receive \(u_i=0\). This aggregate is used in the base support \(L_j\) and fallback weights,
whereas \(q_i(z)\) enters the adaptive diagnostic \(t_i\). Quality-Agent
failure or an omitted candidate ranking falls back to trust \(\tau_i=u_i\).
Trust is therefore an agent assessment or its rule-based substitute, not an
independently calibrated probability.

The base support in Eq.~\eqref{eq:cluster_support} uses
\begin{equation}
\begin{aligned}
\beta&=(0.32,0.22,0.18,0.10,0.06,0.12)^\top,\\
\rho(n_j)&=\min(1,n_j/4),\\
a_j&=\max\!\left(0,2\left[\max_{c_i\in G_j}p_i-0.5\right]\right),
\end{aligned}
\end{equation}
where \(p_i\) is the parsed Scene-Agent method prior, clipped to \([0,1]\)
and defaulting to \(0.5\). The maximum terms retain a strong candidate's
evidence even in a small group, while the means summarize support across
members. Membership contributes a capped term, and the neutral scene prior
adds no positive scene support. With these coefficients, trust contributes
more to the base score than either raw quality or the remaining support
terms. This is an explicit weighting choice; it does not establish that
agent trust is better calibrated than signal quality. The vectors
\(\gamma\) and \(\beta\), like the scene-modulation coefficients, are
part of the experimentally selected configuration and remain fixed during
inference.

If the two largest base supports differ by less than \(0.08\), the Judge
and unavailable-Judge fallback receive
\begin{equation}
\begin{aligned}
T_j&=0.65\bar t_j+0.35t_j^{\max},\\
\widetilde L_j&=L_j+0.06\operatorname{clip}_{[0,1]}(2T_j-1).
\end{aligned}
\end{equation}
Otherwise \(\widetilde L_j=L_j\). PCHV support and harmonic checks use
\(L_j\), without this boost. Fallback ranks
\((\widetilde L_j,\tau_j^{\max},n_j)\) lexicographically; residual ties
retain construction order. Cluster centers are serialized to two decimal
places and evidence scores to four; comparisons use these serialized fields.
For a decision selecting one group, a base-support gap of at least \(0.16\)
from the best group triggers replacement by that best group; a gap in
\([0.10,0.16)\) records a warning without changing the selected group.
Empty or cross-cluster selections use the best base-support group as the
initial repair target. The subsequent harmonic check can override this
target as specified below.

\paragraph{Harmonic configurations and anchors.}
The detector tests ratios of serialized positive centers. Its absolute
ratio tolerance is \(0.3\): \([1.7,2.3]\) for order 2 and \([2.7,3.3]\)
for order 3. A pair is suppressed when another cluster has strictly more
members than both. For order 2, the recommended anchor is the high-rate
group when \(n_{\rm high}\geq2n_{\rm low}\), and the low-rate group
otherwise. For order 3, define counts over all valid candidates:
\begin{equation}
\begin{aligned}
N_{\rm low}&=|\{i:|h_i-\mu_{\rm low}|\leq10~\mathrm{bpm}\}|,\\
N_{\rm high}&=|\{i:h_i\geq1.8\mu_{\rm low}\}|.
\end{aligned}
\end{equation}
The anchor is high if \(N_{\rm high}>N_{\rm low}\), and low otherwise.
These are support heuristics, not proof of a true fundamental frequency.
For a decision selecting one group, PCHV visits pairs in ascending-HR
construction order. The first pair selecting the non-recommended side with
\(L_{\rm anchor}+0.08\geq L_{\rm selected}\) sets the repair target,
overriding an earlier low-support target. Cross-cluster decisions instead
use the base-support repair target.

The archived helper at Git \texttt{4f2f3fb} enables orders \(\{2,3\}\);
the later \texttt{66fc987} helper defaults to \(\{2\}\), as used by the
controlled verifier replay. The available historical main-table summaries
do not establish the enabled order set for every run. We therefore do not
attribute either configuration uniformly to all historical results.

\paragraph{Branch semantics and exceptional inputs.}
For single selection, the effective set contains the requested valid primary
candidate; weighted fusion uses positive-weight, non-excluded valid
candidates; median and trimmed mean use the non-excluded valid set.
Candidate identifiers are normalized as \texttt{METHOD@ROI}. Parsing drops
nonnumeric and nonpositive weights. The verifier clears weights for a
non-weighted strategy and removes weights assigned to excluded candidates;
these adjustments trigger repair. An empty or cross-cluster effective set
also triggers repair. Acceptance preserves the proposed strategy and
normalizes its surviving positive weights. Repair selects the full target
group and excludes all other valid candidates, using the trust floor
\(\eta=0.05\); fallback uses \(\max(\tau_i,u_i,\eta)\). A singleton
returns its HR without weighting.

Without cluster evidence, the archived verifier passes a decision through
without the single-cluster guarantee, while unavailable-Judge fallback uses
the median of valid candidates. With no valid candidates, the fuser returns
a nonfinite result to be counted as a failure. The archived weight parser
does not provide an explicit finiteness guard, so nonfinite weights are
outside the finite-weight guarantees in Appendix~\ref{app:design_properties}.
Deterministic execution here describes replay with fixed evidence and
decisions; it neither removes upstream MLLM dependence nor validates trust
or cluster correctness.

\section{\reviewchange{Limitations}}
\label{app:limitations}

\reviewchange{PhysAgent depends on candidate coverage and evidence quality: inaccurate scene or trust assessments, or a pool missing the true HR, can lead to a wrong hypothesis. Agent calls add inference cost. Future work will study evidence calibration, candidate expansion, and conflict-adaptive inference budgets.}

\end{document}